\documentclass{article}

\usepackage[final]{neurips_2025}
\usepackage[pagebackref=true,breaklinks=true,letterpaper=true,colorlinks,bookmarks=false]{hyperref}
\usepackage[utf8]{inputenc} %
\usepackage[T1]{fontenc}    %
\usepackage{url}            %
\usepackage{booktabs}       %
\usepackage{amsfonts}       %
\usepackage{nicefrac}       %
\usepackage{microtype}      %
\usepackage{xcolor}         %

\usepackage{xspace}
\usepackage{algorithm}
\usepackage{comment}
\usepackage{algorithmicx}
\usepackage{algpseudocode}
\usepackage{amsmath}
\usepackage{cleveref}
\usepackage{xcolor}
\usepackage{colortbl}
\usepackage{listings}
\usepackage{booktabs}
\usepackage{multirow}
\usepackage{pifont}
\usepackage{graphicx}
\usepackage{wrapfig}
\usepackage{placeins}
\usepackage{makecell}
\usepackage{tablefootnote}
\usepackage{enumitem}
\usepackage[normalem]{ulem}
\usepackage{bbm}
\usepackage{dsfont}

\definecolor{pinkred}{RGB}{255, 182, 193}  %
\definecolor{lightgreen}{RGB}{200, 255, 200}  %

\newcommand{\cmark}{\ding{51}}%
\newcommand{\xmark}{\ding{55}}%

\definecolor{commentcolor}{RGB}{115,162,162} %
\ifx \submission \undefined
	\newcommand{\ze}[1]{\textcolor{cyan}{Ze: #1}}

	\newcommand{\joyce}[1]{{\color{brown}{Joyce: #1}}}
	
	\newcommand{\siva}[1]{{\color{magenta}{Siva: #1}}}
	\newcommand{\raquel}[1]{{\color{red}{Raquel: #1}}}
	\newcommand{\ab}[1]{\textcolor{Periwinkle}{#1}}
	\newcommand{\andrei}[1]{\ab{Andrei: #1}}
	
	\newcommand{\jingkang}[1]{{\color{blue}{Jingkang: #1}}}
    
	\newcommand{\yun}[1]{{\color{teal}{yun: #1}}}
	
	\newcommand{\Lily}[1]{{\color{orange}{lily: #1}}}
	
\else
	\newcommand{\ze}[1]{}

	\newcommand{\joyce}[1]{}
	\newcommand{\siva}[1]{{}}
	\newcommand{\raquel}[1]{{}}
	\newcommand{\andrei}[1]{{}}
	\newcommand{\ab}[1]{{#1}}
	
	\newcommand{\jingkang}[1]{{}}
    
	\newcommand{\yun}[1]{{}}
	
	\newcommand{\Lily}[1]{{}}
	
\fi

\newcommand{\name}{\textit{Flux4D}\xspace}
\title{\name: Flow-based Unsupervised 4D Reconstruction}

\author{
	Jingkang Wang$^{1,2}\thanks{Equal contributions.}$ \quad Henry Che$^{1,3*}$\thanks{Work done while a research intern at Waabi.} \quad Yun Chen$^{1,2*}$  \quad Ze Yang$^{1,2}$ \\
	\textbf{Lily Goli}$^{1,2\dag}$ \quad
	\textbf{Sivabalan Manivasagam}$^{1,2}$ \quad \textbf{Raquel Urtasun}$^{1,2}$ \vspace{0.05in}\\
	Waabi$^{1}$ \quad University of Toronto$^{2}$ \quad UIUC$^{3}$  \\
	\url{https://waabi.ai/flux4d}
}

\begin{document}

\maketitle
\vspace{-0.1in}

\begin{abstract}
 Reconstructing large-scale dynamic scenes from visual observations is a fundamental challenge in computer vision. %
 While recent differentiable rendering methods such as NeRF and 3DGS
 have achieved impressive photorealistic reconstruction, they suffer from scalability limitations and require annotations to decouple moving actors from the static scene, such as in autonomous driving scenarios.
 Existing self-supervised methods attempt to eliminate explicit annotations by leveraging motion cues and geometric priors, yet they remain constrained by per-scene optimization and sensitivity to hyperparameter tuning.
 In this paper, we introduce \textit{Flux4D}, a simple and scalable framework for 4D reconstruction of large-scale dynamic driving scenes.
 \textit{Flux4D} directly predicts 3D Gaussians and their motion dynamics to reconstruct sensor observations in a fully unsupervised manner. By adopting only photometric losses and enforcing an ``as static as possible'' regularization, \textit{Flux4D} learns to decompose dynamic elements directly from raw data without requiring pre-trained supervised models or foundational priors simply by training across many scenes. Our approach enables efficient
reconstruction of dynamic scenes within seconds, scales effectively to large datasets, and generalizes well to unseen environments, including rare and unknown objects. Experiments on outdoor driving datasets show \textit{Flux4D} significantly outperforms existing methods in scalability, generalization, and reconstruction quality.

\end{abstract}

\vspace{-0.1in}
\section{Introduction}
\label{sec:intro}

Reconstructing the 4D physical world from visual observations captured in the wild is a key goal in computer vision, with applications in virtual reality and robotics, including autonomous driving.
High-quality reconstructions provide the foundation for scalable simulation environments that enable safer and more efficient autonomy development.
Unlike artist-created environments, environments built automatically with data collected by sensor-equipped vehicles are more realistic, are more cost-efficient, and capture the diversity of the real world~\cite{wang2021advsim,unisim,manivasagam2023towards}.

Advances in differentiable rendering approaches such as
Neural Radiance Field (NeRF)~\cite{mildenhall2020nerf} and 3D Gaussian Splatting (3DGS)~\cite{3dgs} have enabled high-quality reconstruction of dynamic scenes~\cite{unisim,yan2024street,zhou2024drivinggaussian,tonderski2024neurad,khan2024autosplat,chen2025salf,turki2025simuli}.
These methods decompose scenes into a static background and a set of dynamic actors
using human annotations such as 3D tracklets or dynamic masks, and then perform rendering on the composed representation, optimizing to reconstruct the input observations.
While they achieve impressive visual fidelity, their reliance on manual annotations to decompose static and dynamic elements increases costs and time,
preventing these methods from scaling to large sets of unlabelled data.
Some approaches leverage pre-trained perception models to generate annotations automatically,
but this can cause artifacts when the model predictions are noisy or incorrect, which can be difficult to recover from during reconstruction.
Moreover, these methods typically require hours to reconstruct each scene on consumer GPUs.
These two main issues, expensive annotation costs
and slow per-scene optimization, limit the scalability of these methods.

Recent works have explored self-supervised approaches to eliminate the reliance on human
annotations and learn the decomposition of static and dynamic actors directly from data.
This is a challenging task due to the ambiguity of actor motion over time, coupled with
spatial geometry and appearance variations.
One strategy attempts to improve the decomposition by incorporating additional regularization terms such as geometric constraints \cite{peng2024desire} or cycle consistency \cite{yang2023emernerf}, or performing multi-stage training \cite{huang2024s3}.
Another strategy is to leverage foundation models for additional semantic features or priors
\cite{peng2024desire, chen2023periodic, yang2023emernerf}.
However, the resulting complex models can be sensitive to hyperparameters, slow to train, and unable to generalize to new scenes.
Moreover, they often have poor decomposition results, and struggle to render novel views, limiting their usability.

As an alternative to costly per-scene optimization, generalizable approaches~\cite{chen2021mvsnerf,wang2021ibrnet,charatan2023pixelsplat,chen2024mvsplat,hong2024lrm,wei2024meshlrm,zhang2025gs} use feed-forward neural networks to predict
scene representations directly from observations, enabling efficient reconstruction within seconds.
However, these approaches are designed for small-scale environments, can only process a few low-resolution images (typically 1-4 views with resolutions below 512px), and
primarily focus on static scenes~\cite{charatan2023pixelsplat,chen2024mvsplat} or only
dynamic objects \cite{ren2025l4gm}.
When handling large scenes with many dynamic elements, they rely on costly annotations \cite{chen2025g3r, ren2024scube}, limiting their scalability. %
Most recently, DrivingRecon~\cite{lu2024drivingrecon} and STORM~\cite{yang2025storm} propose feed-forward, self-supervised approaches for driving scenes.
While promising, these methods focus on the sparse reconstruction setting and can only handle a small number ($\leq12$) of low-resolution ($\leq360$px) input views  before reaching compute limits,
and still depend on pre-trained vision models for semantic guidance, constraining their fidelity, scalability and applicability to downstream simulation.

\begin{figure}[t]
	\centering
    \includegraphics[width=\columnwidth]{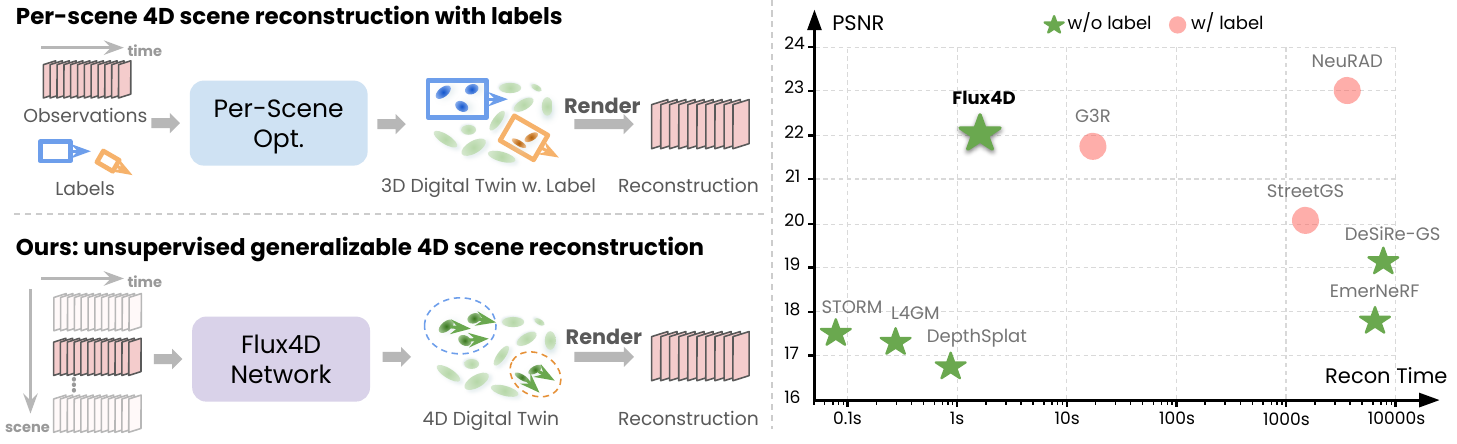}
    \vspace{-0.23in}
    \caption{
\textbf{\name is a simple and scalable framework for unsupervised 4D reconstruction. }%
            \textbf{Left:} Paradigms for 4D reconstruction.
            \textbf{Right:} realism-speed comparisons with existing works.
    }
	\label{fig:paradigm}
    \vspace{-0.16in}
\end{figure}

In this paper, we propose \name, an \textit{unsupervised} and \textit{generalizable} reconstruction approach that enables accurate and efficient 4D driving scene reconstruction at scale.
Without any annotations, \name
predicts 3D Gaussians along with motion parameters directly in 3D space from multi-sensor observations within seconds, enabling efficient scene reconstruction.
Our reconstruction paradigm is illustrated
in Fig.~\ref{fig:paradigm}.
\name uses a remarkably minimalist design that employs only photometric losses and a simple static-preference prior, without requiring complex regularization schemes or external supervision to learn the motion that prior works leverage.
We find that the key ingredient for \name to accurately recover geometry, appearance, and motion flow comes from learning across a diverse range of scenes.
Moreover, \name's use of LiDAR data, commonly available in the autonomous driving domain, enable handling of a large number ($\geq60$) of high-resolution (1080px) input multi-view images, achieving high-fidelity reconstruction and scalable simulation.
Our 3D design yields a compact and geometrically consistent representation across views, improving efficiency, enabling explicit multi-view flow reasoning and reducing appearance-motion ambiguity.

Experiments on outdoor driving datasets PandaSet \cite{xiao2021pandaset} and WOD \cite{waymo}  demonstrate that \name achieves better scene
decomposition and novel view synthesis than previous state-of-the-art annotation-free reconstruction methods, and is competitive with per-scene optimization methods that use human annotations.
We also show that \name{} can be trained to %
predict sensor observations in future frames, akin to next-token prediction, but applied to dynamic 3D scenes.
Finally, we showcase using \name's reconstruction for controllable camera simulation via scene editing and novel view rendering at high resolution ($\geq$ 1080px).
\name highlights the power of unsupervised learning for 4D scene reconstruction, enabling efficient scaling to vast unlabeled datasets.

\vspace{-0.12in}
\section{Related Work}
\vspace{-0.05in}
\label{sec:related}

\vspace{-0.07in}
\paragraph{Optimization-based 4D reconstruction:}
Inspired by differentiable rendering~\cite{mildenhall2020nerf,3dgs}, recent approaches use deformation fields~\cite{pumarola2021d,park2020deformable,yang2024deformable,wu20244d} to model dynamic scenes but still struggle with real-world complexity due to overparameterization and poor static-dynamic decomposition. While some methods address this by using human annotations (3D tracklets, semantic models) to explicitly separate static~\cite{wang2022cadsim,yang2023reconstructing} and dynamic elements~\cite{ost2021neural,unisim,liu2023real,turki2023suds,pun2023lightsim,yan2024street,chen2024omnire,hess2024splatad,yang2025genassets}, they remain limited by annotation quality and availability. Self-supervised alternatives using motion cues and physics-informed priors~\cite{wu2022d,yang2023emernerf,chen2023periodic,huang2024s3,peng2024desire} reduce dependence on annotations but typically require complex regularization schemes and expensive per-scene optimization. In contrast, our approach reconstructs dynamic 4D scenes without explicit supervision or per-scene optimization, achieving scalable reconstruction through simple photometric losses with minimal regularization.

\vspace{-0.12in}
\paragraph{Generalizable reconstruction:}
Generalizable methods infer scene representations directly from observations without per-scene optimization~\cite{chen2021mvsnerf,wang2021ibrnet,charatan2023pixelsplat,chen2024mvsplat,hong2024lrm,wei2024meshlrm,zhang2025gs}, leveraging large training datasets to improve reconstruction quality in novel environments. However, existing approaches primarily target static scenes, struggling with dynamic environments due to computational constraints and dependence on sparse, low-resolution inputs. Recent advances attempt to overcome these limitations using efficient architectures~\cite{ziwen2024long} or iterative refinement~\cite{chen2025g3r}, but still rely on 3D annotations. In contrast, \name generalizes to unseen dynamic scenes by predicting 3D Gaussians with their motion directly from raw observations without external supervision.

\vspace{-0.1in}
\paragraph{Unsupervised world models:}
Our work relates to recent advances in unsupervised world models, which learn predictive representations of environments without explicit supervision. These approaches typically tokenize visual data into discrete or continuous representations~\cite{hu2023gaia,gao2024vista,wang2024drivedreamer,zheng2024occworld,min2024driveworld} processed by autoregressive or diffusion-based models to predict future states. While demonstrating impressive visual quality, such methods generally lack interpretable 3D structure, limiting precise control over generated content. Existing solutions often produce lower-resolution outputs with reduced temporal consistency, are typically restricted to single modalities (\textit{e.g.}, camera~\cite{hu2023gaia,gao2024vista,li2024drivingdiffusion} or LiDAR~\cite{khurana2023point,zhang2023learning,yang2024visual,agro2024uno}), and require substantial computational resources.
While our primary focus is reconstruction, \name's ability to simultaneously model motion dynamics and predict future frames shares conceptual similarities with world models. Unlike these approaches, \name uses explicit 3D representation, providing 3D interpretability, controllability and spatiotemporal consistency.

\vspace{-0.1in}
\paragraph{Unsupervised generalizable reconstruction:}
Most recently, DrivingRecon~\cite{lu2024drivingrecon} and STORM~\cite{yang2025storm} explore
unsupervised generalizable 4D reconstruction for driving scenes, using feed-forward networks to predict the velocities of 3D Gaussians.
Despite impressive performance, they can process only sparse (3-4), low-resolution ($\leq256\times512$) frames with substantial computational requirements and rely on pre-trained vision models (DeepLabv3+~\cite{chen2017rethinking}, SAM~\cite{kirillov2023segment}, ViT-Adapter~\cite{chen2023vision}) for additional supervision, limiting their scalability and applicability.
\name achieves better performance with a simpler and more scalable approach, and through our novel incorporation of LiDAR to initialize the scene, can handle full HD images with denser views ($> 60$) while being computationally efficient.
Please see Appendix~\ref{sec:discussion} for more discussions.

\section{Scalable 4D Reconstruction with \name}
\label{sec:method}

\begin{figure*}[t]
	\centering
	\includegraphics[width=1.0\textwidth]{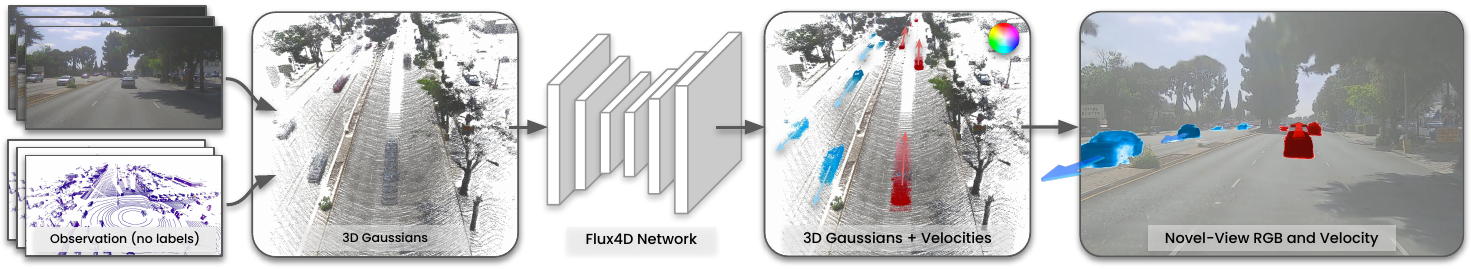}
	\vspace{-0.21in}
    \caption{\textbf{Model overview.} \textit{\name} reconstructs 4D world by predicting 3D Gaussians with velocities 
given unlabelled sensor observations, and trained with the photometric reconstruction objective. The resultant model can be used for RGB and flow synthesis from novel views.}
	\label{fig:method}
\end{figure*}

Given a sequence of camera and LiDAR data captured by a robot sensor platform, we aim to reconstruct the underlying 4D scene representation that disentangles static and dynamic entities and supports high-quality rendering at novel viewpoints.
Such a representation can enable future prediction and counterfactual simulation.
To achieve scalable 4D scene reconstruction, our method should be unsupervised, meaning it uses no annotations, and fast, running in seconds.
Towards this goal, we propose \name, an unsupervised and generalizable approach that learns to reconstruct 4D scenes via three simple steps (Fig.~\ref{fig:method}).
We first lift the sensor observations at each timestep to a set of initial 3D Gaussians. 
We then feed the initial representation to a network to predict 3D flow and refined attributes for each 3D Gaussian. 
Finally, we supervise the network solely through reconstruction and static-preference losses.

\subsection{Scene Representation}
\label{sec:representation}
Our approach takes a set of posed camera images $\mathcal{I} = \{\mathbf{I}_k\}_{1 \leq k \leq K}$ 
and LiDAR point clouds $\mathcal{P} = \{\mathbf{P}_k \}_{1 \leq k \leq K}$ 
captured over time
by a moving platform and outputs a scene representation with geometry, appearance, and 3D flow.
We represent the scene using a set of 3D Gaussians
$\mathcal{G} =\{\mathbf g_i\}_{1\leq i \leq M}$.
Each Gaussian point $g_i$ is parameterized by its center position $\mathbf{p}_i$ ($\mathbb R^3$), scale ($\mathbb R^3$), orientation ($\mathbb R^4$), color ($\mathbb R^3$) and opacity ($\mathbb R^1$) ~\cite{3dgs}.
Additionally, we augment each Gaussian with a learnable instantaneous velocity $\mathbf{v}_i \in \mathbb{R}^3$ and a fixed capture time $t_i$. 
We denote the sets of velocities and timestamps for all Gaussians as $\mathcal{V} = \{\mathbf{v}_i\}_{1\leq i \leq M}$ and $\mathcal{T} = \{t_i \}_{1\leq i \leq M}$.

\paragraph{Initialization:}
We initialize Gaussian positions from LiDAR points $\mathbf{P}_k$ from each source frame in the sequence, set scales based on the average distance to nearby points, and assign colors by projecting these points onto the corresponding camera image $\mathbf{I}_k$.
Each Gaussian's timestamp $t_i$ is assigned
the capture time of its source LiDAR frame, and velocities are initialized to zero.
We aggregate source frame Gaussians to create $\mathcal{G}_{\mathrm{init}}$.

\subsection{Predicting Flow and Rendering}
\label{sec:rendering}
Inspired by recent advances in 4D reconstruction~\cite{wu2022d,yang2023emernerf,peng2024desire,zhang2024visionpad,lu2024drivingrecon,yang2025storm}, we propose to learn a time-dependent velocity field to model the dynamics of driving scenes. Given the initial velocity-augmented Gaussians $\mathcal{G}_{\mathrm{init}}$, we leverage a neural reconstruction function $f_\theta$
that outputs the refined Gaussian parameters $\mathcal{G}$ and the predicted velocities $\mathcal{V}$:
\begin{align}
	\mathcal{G}, \mathcal{V}  = f_\theta(\mathcal{G}_{\mathrm{init}}, \mathcal{T}).
\end{align}
With the predicted velocities $\mathcal{V}$, each Gaussian can be propagated from its initial timestep $t_i$ to any target timestep $t'$ using a linear motion model:
\begin{align}
	\mathbf p_i^{t'} = \mathbf p_i^{t_i} + \mathbf{v}_i \cdot (t' - t_i),
	\label{eq:movement}
\end{align}
where $\mathbf p_i^{t'}$ is the Gaussian position at time $t'$, $\mathbf{v}_i$ and ${t}_i$ are its velocity and capture time.
This formulation enables continuous, temporally consistent reconstruction under a constant velocity assumption.
We find this simple motion model can already achieve reasonable performance when reconstructing outdoor driving scenes with short time horizons ($\sim1s$), an observation aligned with existing works~\cite{peng2024desire,lu2024drivingrecon,li2025gvfi,yang2025storm}. Moreover, we investigate higher-order polynomial motion models, as discussed in Sec.~\ref{sec:enhancement} and Table~\ref{tab:ablation}.

\subsection{Unsupervised Learning of Dynamics}
\label{sec:learning}
We now describe how the method learns to disentangle the scene dynamics.
The network $f_\theta$ is trained in a fully self-supervised manner, without requiring explicit 3D annotations.
Given the predicted Gaussians $\mathcal{G}$, we move the Gaussians to target time $t^\prime$ using Eqn.~(\ref{eq:movement}), render the scene using differentiable rasterization~\cite{3dgs} to generate color and depth images, and compare them against the real sensor observations $\mathcal I$ and $\mathcal P$.
To prevent unnecessary motion and encourage stability, we introduce an ``\textit{as static as possible}'' regularization. The total loss $\mathcal{L}$ is defined as:
\begin{align}
	\mathcal{L} =  \mathcal{L}_\text{recon} + \lambda_\text{vel} \mathcal{L}_\text{vel},
	\label{eq:loss}
\end{align}
where $\mathcal{L}_\text{recon}$ represents the
reconstruction loss, consisting of $L_1$ and structural similarity losses w.r.t the images, and an $L_1$ depth loss in the image plane compared to the projected LiDAR,
and $\mathcal{L}_\text{vel}$ serves as a velocity regularization term that minimizes motion magnitudes:
\begin{align}
	\mathcal{L}_\text{recon} = \lambda_\text{rgb} \mathcal{L}_\text{rgb} + \lambda_\text{SSIM} \mathcal{L}_\text{SSIM} + \lambda_\text{depth} \mathcal{L}_\text{depth},
\end{align}
\begin{align}
	\mathcal{L}_\text{vel} = \frac{1}{M} \sum_{i} \| \mathbf{v}_i \|_2.
\end{align}
We train $f_\theta$ across a diverse set of scenes.
Notably, we find that training across many scenes enables the network to \textit{automatically} decompose static and dynamic components in urban scenes 
without requiring the complex regularizations used in prior per-scene optimization techniques
~\cite{wu2022d,yang2023emernerf,chen2023periodic,huang2024s3,peng2024desire}.
This highlights the effectiveness of data-driven priors as a powerful form of implicit regularization and the scalability of this simple framework.

\subsection{Improving Realism and Flow}
\label{sec:enhancement}

The aforementioned components form the core of our approach, termed \name-base.
\name-base can already disentangle motion and render novel views with high quality.
We further improve \name-base through two enhancements that further recover more fine-grained appearance and refined flow, resulting in our final model, \name.

\vspace{-0.05in}
\paragraph{Iterative refinement:}
\name-base recovers the overall scene appearance, but often lacks fine-grained details.
We hypothesize that this limitation stems from the constrained capacity of a single-step feedforward network, and imperfect initialization due to occlusions.
To mitigate this, we introduce an iterative refinement mechanism inspired by G3R~\cite{chen2025g3r}, leveraging 3D gradients as feedback to enhance reconstruction quality.
Specifically, after each forward pass and generation of rendered color and depth at the supervision views, we compute the 3D gradients of the Gaussians according to the loss function Eqn.~(\ref{eq:loss}),
and provide the generated Gaussians and gradients as input to a network $f_\phi$ to further refine them.
This process progressively corrects color inconsistencies and sharpens details within as few as two iterations. By incorporating iterative feedback, our method achieves higher-fidelity reconstruction, particularly in regions with complex appearance variations, while preserving the efficiency and scalability of \name-base. 

\vspace{-0.05in}
\paragraph{Motion enhancement:}

\name-base recovers the overall scene flow accurately (Table~\ref{tab:ablation}).
We further introduce \textit{polynomial motion} parameterizations to better model actor behaviors like acceleration, braking or turning. Please see Appendix~\ref{sec:motion_details} for more details and comparisons.
Exploring more advanced velocity models~\cite{li2025gvfi} or implicit flow representations is an exciting direction for future work.
To further improve the flow and appearance quality of dynamic actors, we modify the loss function to focus on dynamic regions.
Specifically, we render the flow in the image plane and apply pixel-wise re-weighting to the photometric loss.
This gives higher importance to faster-moving regions during training, which typically occupy fewer pixels and would contribute less to the overall loss.

\section{Experiments}
\label{sec:experiments}

\begin{table*}[t]
	\centering
	\caption{\textbf{Comparison to SoTA unsupervised methods on novel view synthesis.}
		We evaluate photorealism, geometry, and speed metrics against per-scene optimization methods and generalizable methods.
		$^\dagger$~denotes the need for pre-trained vision models.
		\name{} surpasses unsupervised and achieves competitive performance with supervised methods (top block), without requiring 3D labels.}
	\label{tab:comparison}
	\resizebox{\textwidth}{!}{
		\begin{tabular}{lccccccccccc}
			\toprule
			\multirow{2}{*}{\textbf{Methods}} & \multirow{2}{*}{\textbf{\textit{Unsup.}}} & \multirow{2}{*}{\textbf{\textit{Gen.}}} & \multicolumn{4}{c}{\textbf{Dynamic-only}} & \multicolumn{4}{c}{\textbf{Full image}} & \multicolumn{1}{c}{\textbf{Recon speed}} \\
			& & & PSNR$\uparrow$ & SSIM$\uparrow$ & $D_{\mathrm{MAE}}\downarrow$ & $V_{\mathrm{RMSE}}\downarrow$ & PSNR$\uparrow$ & SSIM$\uparrow$ & $D_{\mathrm{MAE}}\downarrow$ & $V_{\mathrm{RMSE}}\downarrow$ & Time↓ \\
			\midrule
			\multicolumn{3}{l}{\textbf{\textit{Recon. with labels (reference)}}} \\
			\textcolor{gray}{NeuRAD~\cite{tonderski2024neurad}} & \textcolor{gray}{\xmark} & \textcolor{gray}{\xmark} &  \textcolor{gray}{23.01} &	\textcolor{gray}{0.734} &	\textcolor{gray}{1.98} &  \textcolor{gray}{--} &	\textcolor{gray}{24.61}	& \textcolor{gray}{0.685}	& \textcolor{gray}{2.30} & \textcolor{gray}{--} & \textcolor{gray}{$\sim$60min} \\
			\textcolor{gray}{StreetGS~\cite{yan2024street}} & \textcolor{gray}{\xmark} & \textcolor{gray}{\xmark} & \textcolor{gray}{20.06}  &  \textcolor{gray}{0.605} & \textcolor{gray}{1.02} &  \textcolor{gray}{--} & \textcolor{gray}{23.38} &  \textcolor{gray}{0.680} & \textcolor{gray}{0.84} &  \textcolor{gray}{--} & \textcolor{gray}{$\sim$28min}  \\
			\textcolor{gray}{G3R~\cite{chen2025g3r}} & \textcolor{gray}{\xmark} & \textcolor{gray}{\cmark}  & \textcolor{gray}{21.85} &	\textcolor{gray}{0.670} &	\textcolor{gray}{2.33} &	\textcolor{gray}{--} & \textcolor{gray}{24.35}	 & \textcolor{gray}{0.686} &	\textcolor{gray}{1.96} & \textcolor{gray}{--} & \textcolor{gray}{17s} \\
			\midrule
			\multicolumn{3}{l}{\textbf{\textit{Unsupervised recon.}}} \\
			EmerNeRF$^\dagger$~\cite{yang2023emernerf} & \cmark & \xmark & 17.79	 & 0.411 &	6.09
			& 0.318 & 22.80 &	0.624 & 4.24  & 0.432 & $\sim$100min  \\
			DeSiRe-GS$^\dagger$~\cite{peng2024desire} & \cmark & \xmark & 19.08 &	0.477 &	3.36 & 0.297  & 22.25 &	 0.608 & 24.89 & 0.322 & $\sim$120min  \\
			DepthSplat$^*$~\cite{xu2024depthsplat}  & \cmark &   \cmark & 16.87 &  0.425 & 6.18 & --  & 21.40 &  0.595 & 2.73 & -- & 0.87s \\
			L4GM~\cite{ren2025l4gm} & \cmark &  \cmark &  17.36 & 0.343	 &  -- &  -- & 19.38 & 0.465 &  -- & -- & {0.32s}  \\
			STORM$^\dagger$~\cite{yang2025storm} & \cmark &  \cmark & 17.65 &	0.367 &	5.24 &	0.203 & 20.79 &	0.508 &	4.80	& 0.238 & \textbf{0.08s} \\
			\textbf{\name-\textit{base} (Ours)} & \cmark &  \cmark &
			18.62 &	0.454 & 2.06 & 0.181	& 20.88	& 0.575	& 1.37	& 0.211 & 0.13s
			 \\
			\textbf{\name (Ours)} & \cmark &  \cmark & \textbf{21.99}  & \textbf{0.662} &  \textbf{1.63} & \textbf{0.157}  & \textbf{23.84 }&  \textbf{0.675} & \textbf{1.07} & \textbf{0.182} & 1.8s \\
			\bottomrule
		\end{tabular}
	}
	\vspace{-0.1in}
\end{table*}

\begin{figure*}[t]
	\centering
	\vspace{0.1in}
	\includegraphics[width=1.0\textwidth]{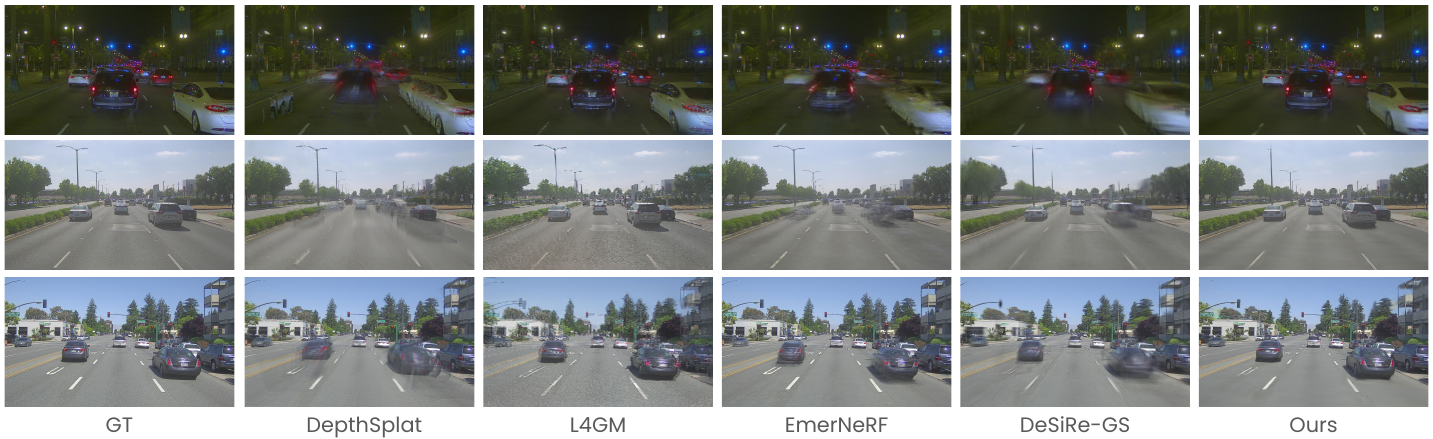}
	\vspace{-0.2in}
	\caption{\textbf{Qualitative results for NVS on PandaSet}. Rendered RGB images from novel views show that our method achieves better image quality across a variety of urban scenes,
	 with
	 crisper edges and sharper
	 dynamic actors compared to baselines.}
	\label{fig:comp_1s}
	\vspace{-0.05in}
\end{figure*}

We evaluate \name against the current state-of-the-art (SoTA) self-supervised scene reconstruction methods, including both per-scene optimization and generalizable approaches.
We also report the performance of supervised methods that do require annotations to model dynamics as a reference.
We perform experiments on multiple outdoor dynamic datasets and assess novel view appearance and depth, as well as recovered flow.
We also ablate \name's design and show that \name scales with more data.
Finally, we demonstrate the controllability of our predicted scene representation for realistic camera simulation.

\subsection{Experimental Details}
\label{subsec:exp_details}

\begin{figure*}[t]
	\centering
	\includegraphics[width=1.0\textwidth]{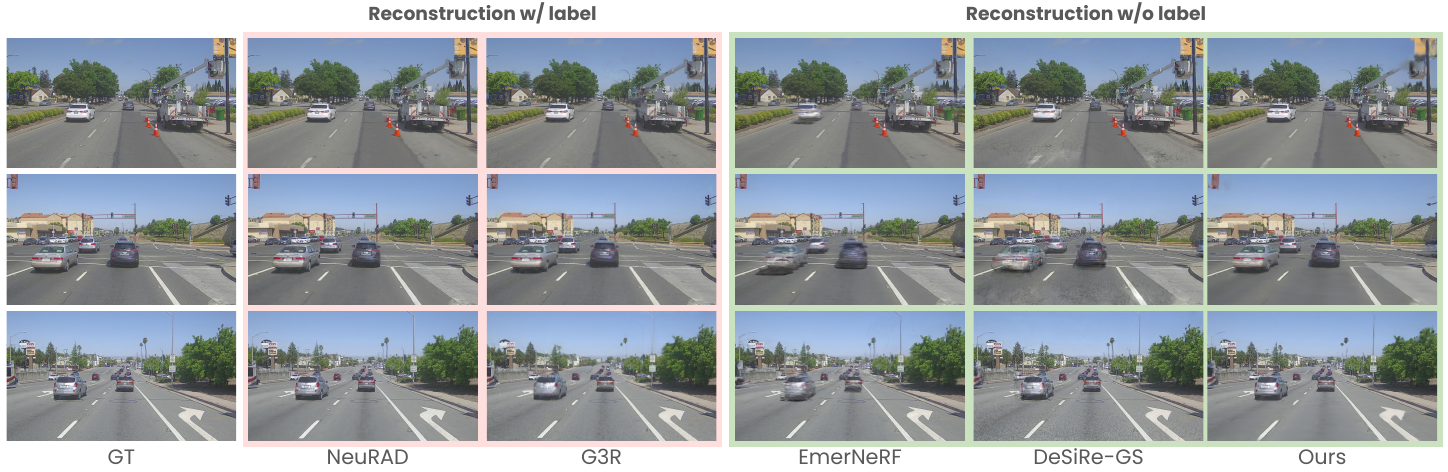}
	\vspace{-0.2in}
	\caption{\textbf{NVS on longer-horizon logs.} Qualitative comparison shows that our method outperforms SoTA
	unsupervised baselines,
	 by maintaining better estimation of actor movements in longer horizon. We shrink the gap in quality to {supervised methods.}} %
	\vspace{-0.05in}
	\label{fig:comp_8s}
\end{figure*}

\begin{figure*}[t]
	\centering
	\includegraphics[width=1.0\textwidth]{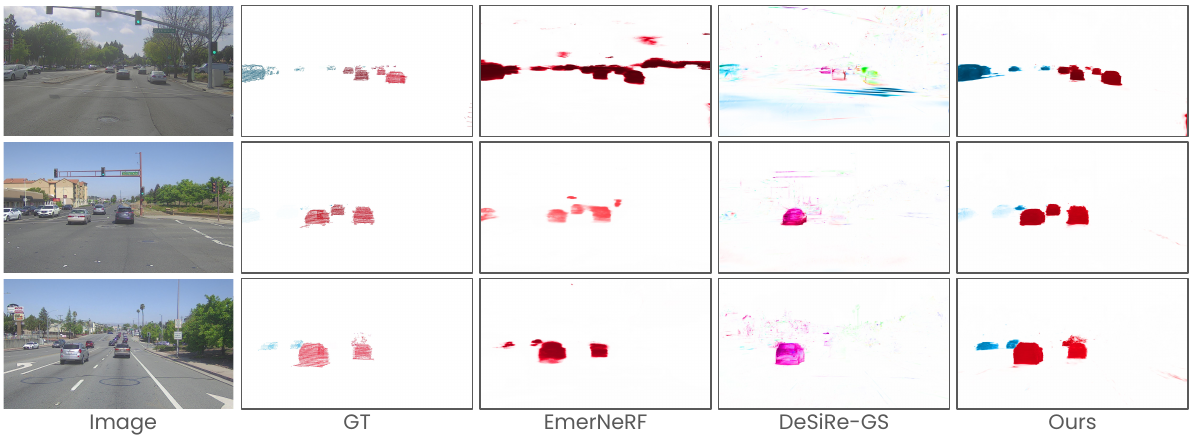}
	\vspace{-0.2in}
	\caption{\textbf{Estimating motion flows.} We compare our estimated motion with prior unsupervised methods through rendered flow, showing accurate static region detection and sharper actor flow edges.
	}
	\label{fig:comp_flow}
	\vspace{-0.03in}
\end{figure*}

\paragraph{Experiment setup:}
We conduct experiments on outdoor driving scenes from PandaSet~\cite{xiao2021pandaset} and Waymo Open Dataset (WOD)~\cite{waymo}.
From PandaSet's 103 dynamic scenes (1080p cameras, 64-beam LiDARs, 10Hz), we select 10 diverse scenes for validation and use the rest for training.
We use the front camera and 360$^\circ$ LiDAR, both collected at 10 Hz.
To compare against existing feed-forward generalizable reconstruction methods that can only take a small number of frames as input, we report scene reconstruction results on short 1.5s windows within the validation sequences.
Each method takes as input frames 0, 2, 4, 6, 8, 10, and is evaluated
on frames 1, 3, 5, 7, 9 (\emph{interpolation}) and 11-15 (\emph{future prediction}).
We sample a new snippet every 20 frames, yielding four non-overlapping evaluation snippets per log.
We also evaluate against per-scene optimization methods over the full duration of the validation sequence (8 seconds) in the interpolation setting (every other frame is held out).
For WOD evaluation, we follow the NVS setting in DrivingRecon~\cite{lu2024drivingrecon}, using the Waymo-NOTR subset with three front cameras,
taking $\{t-2, t-1, t+1\}$ frames as input, and generating the interpolated frame at time $t$, where $t$ is every tenth frame in each sequence.
Finally, we evaluate scene flow estimation perpformance on PandaSet and WOD (official validation set with 202 logs).
As existing scene flow estimation methods cannot directly predict flows at novel timesteps, we evaluate scene flow on the input frames. We restrict evaluation to LiDAR points within the camera field of view (FoV) following~\cite{yang2025storm}.

\vspace{-0.1in}
\paragraph{Baselines:} We compare against SoTA unsupervised scene reconstruction approaches: (1) \textit{Self-supervised per-scene optimization:} EmerNeRF~\cite{yang2023emernerf} and DeSiRe-GS~\cite{peng2024desire}, which reconstruct dynamic scenes using geometry priors, cycle consistency, and pre-trained vision models (FiT3D~\cite{yue2024fit3d} and DINOv2~\cite{oquab2023dinov2});
(2) \textit{Generalizable methods:} L4GM$^*$~\cite{ren2025l4gm}, a 4D reconstruction model adapted to driving scenes using depth supervision;
DepthSplat$^*$, an extension of~\cite{xu2024depthsplat} that unprojects LiDAR points using estimated depth for 3D Gaussian prediction; DrivingRecon~\cite{lu2024drivingrecon}, which builds a 4D feed-forward model utilizing learned priors from pre-trained vision models (SAM~\cite{kirillov2023segment} and DeepLab-v3~\cite{chen2017rethinking}); and STORM~\cite{yang2025storm} which predicts per-pixel Gaussians and their motion in a feed-forward manner.
For reference, we also include SoTA methods that use ground-truth 3D tracklets: StreetGS~\cite{yan2024street} and NeuRAD~\cite{tonderski2024neurad} (compositional 3DGS/NeRF), as well as G3R~\cite{chen2025g3r} (iterative refinement of compositional 3DGS).
Apart from reconstruction methods, we also compare with representative scene flow estimation methods NSFP~\cite{li2021neural} and FastNSF~\cite{li2023fast} as a reference.

\vspace{-0.1in}
\paragraph{Metrics:}
We report standard metrics to measure the photorealism, geometric and motion accuracy using
PSNR, SSIM, and depth MAE (D$_{\mathrm{MAE}}$) and normalized velocity RMSE (V$_{\mathrm{RMSE}}$). Please see Appendix~\ref{sec:exp_details} for more details. Results are reported on both full images (non-sky regions) and dynamically moving regions.
For scene-flow quality, we report EPE3D, $Acc_5$ and ${Acc}_{10}$ (fraction of points with error $\leq$ 5/10 cm), angular error in radians ($\theta_{\epsilon}$), three-way EPE~\cite{chodosh2024re}: background-static (BS), foreground-static (FS), and foreground-dynamic (FD), bucketed normalized EPE~\cite{khatri2024can}, and inference speed. On WOD, where semantic labels are coarse, we follow EulerFlow~\cite{vedder2025neural} and report bucketed normalized EPE for \emph{Background (incl. Signs)}, \emph{Vehicles}, \emph{Pedestrians}, and \emph{Cyclists} only.

\vspace{-0.1in}
\paragraph{\name implementation details:}
We adopt a 3D U-Net with sparse convolutions~\cite{tangandyang2023torchsparse}
for $f_\theta$.
To handle unbounded scenes, we place random points on a spherical plane at a far distance to model sky and far-away regions.
We also add random points within a 3D sphere
following~\cite{yan2024street} to increase model robustness.
Our model processes full-resolution images ($\geq 1920 \times 1080$) in all experiments and can be efficiently scaled to higher resolutions without significant overhead.
Unless otherwise stated, all models are trained for 30,000 iterations on $4\times$ NVIDIA L40S (48G) GPUs, taking approximately 2 days. %
The reconstruction loss weights $\lambda_{\mathrm{rgb}}, \lambda_{\mathrm{SSIM}}, \lambda_{\mathrm{depth}}$ are set as 0.8, 0.2 and 0.01 respectively.
The velocity regularization weight $\lambda_\mathrm{vel}$ is set as 5e-3.

\subsection{Scalable 4D Reconstruction}

{
\setlength{\tabcolsep}{5pt}
\begin{table}[t]
	\centering
	\caption{\textbf{
		Full sequence reconstruction.} \name outperforms unsupervised methods for
		8-second reconstructions on dynamic regions and full image, closing the gap with supervised methods.}
	\vspace{-0.05in}
	\label{tab:comparison_full_snippet}
	\resizebox{0.9\columnwidth}{!}{
		\begin{tabular}{lcccccccc}
			\toprule
			\multirow{2}{*}{\textbf{Methods}} & \multicolumn{4}{c}{\textbf{Dynamic-only}} & \multicolumn{4}{c}{\textbf{Full image}} \\
			& PSNR$\uparrow$ & SSIM$\uparrow$ & $D_{\mathrm{MAE}}\downarrow$ & $V_{\mathrm{RMSE}}\downarrow$ & PSNR$\uparrow$ & SSIM$\uparrow$ & $D_{\mathrm{MAE}}\downarrow$ & $V_{\mathrm{RMSE}}\downarrow$ \\
			\midrule
			\multicolumn{3}{l}{\textbf{\textit{Recon. with labels (reference)}}} \\
			\textcolor{gray}{NeuRAD~\cite{tonderski2024neurad}} & \textcolor{gray}{22.99}  & \textcolor{gray}{0.719}  & \textcolor{gray}{1.71}  & \textcolor{gray}{--} &  \textcolor{gray}{24.99} & \textcolor{gray}{0.679} &  \textcolor{gray}{2.29} & \textcolor{gray}{--} \\
			\textcolor{gray}{StreetGS~\cite{yan2024street}} & \textcolor{gray}{21.63} & \textcolor{gray}{0.701}  & \textcolor{gray}{0.94}  & \textcolor{gray}{--} & \textcolor{gray}{23.89}  &  \textcolor{gray}{0.708} & \textcolor{gray}{0.87}  & \textcolor{gray}{--} \\
			\textcolor{gray}{G3R~\cite{chen2025g3r}} & \textcolor{gray}{20.60} & \textcolor{gray}{0.573}  & \textcolor{gray}{2.16} & \textcolor{gray}{--} &  \textcolor{gray}{23.15}  &  \textcolor{gray}{0.636} & \textcolor{gray}{2.01}    & \textcolor{gray}{--} \\
			\midrule
			\multicolumn{3}{l}{\textbf{\textit{Unsupervised recon.}}} \\
			EmerNeRF$^\dagger$~\cite{yang2023emernerf} & 18.65 & 0.437 & 4.48 &  0.478 &  23.42 &  0.627 & 3.09 & 0.975 \\
			DeSiRe-GS$^\dagger$~\cite{peng2024desire} & 19.76 & 0.544  & 4.08 & 0.312 &  22.91 &  0.659 & 4.07  & 0.395 \\
			\textbf{\name (Ours)} & \textbf{21.94} &  \textbf{0.658} & \textbf{1.57} & \textbf{0.162} &  \textbf{23.72} &  \textbf{0.670} &  \textbf{1.10} & \textbf{0.186}\\
			\bottomrule
		\end{tabular}
	}
\end{table}
}

\begin{table}[t]
	\centering
	\begin{minipage}{0.59\linewidth}
		\caption{\textbf{NVS on WOD~\cite{waymo}.} We achieve significant improvements over generalizable baselines.}
		\label{tab:drivingrecon}
		\vspace{-0.05in}
		\resizebox{\columnwidth}{!}{
			\begin{tabular}{lccccccc}
				\toprule
				\multirow{2}{*}{\textbf{Methods}}  & \multicolumn{3}{c}{\textbf{Full Image}} & \multicolumn{2}{c}{\textbf{Dynamic}} & \multicolumn{2}{c}{\textbf{Static}} \\
				& PSNR & SSIM & LPIPS & PSNR & SSIM & PSNR & SSIM \\
				\midrule
				LGM~\cite{tang2024lgm}    & 17.49 & 0.47 & 0.33 & 17.79 & 0.49 & 15.37 & 0.39  \\
				PixelSplat~\cite{charatan2023pixelsplat}  & 18.24 &  0.56 & 0.30 & 18.63 & 0.58 & 16.96 & 0.44  \\
				MVSplat~\cite{chen2024mvsplat}  & 19.00 & 0.57 & 0.28 & 19.29 & 0.58 & 17.35 & 0.47   \\
				L4GM~\cite{ren2025l4gm}   &  17.63 & 0.54 & 0.31 & 18.58 & 0.56 & 16.78 & 0.43   \\
				DrivingRecon~\cite{lu2024drivingrecon} & 20.63 & 0.61 & 0.21 & 20.97 & 0.62 & 19.70 & 0.51  \\
				\textbf{\name}      & \textbf{26.62}  & \textbf{0.82}  & \textbf{0.18}  & \textbf{26.86}  & \textbf{0.83}  & \textbf{26.09}  & \textbf{0.80}  \\
				\bottomrule
			\end{tabular}
		}
	\end{minipage}
	\hfill
	\begin{minipage}{0.4\linewidth}
		\centering
		\setlength{\tabcolsep}{3pt}
		\caption{\textbf{Future prediction.} We surpass unsupervised and supervised methods.}
		\label{tab:future_prediction}
		\vspace{-0.05in}
		\resizebox{\columnwidth}{!}{
			\begin{tabular}{lcccc}
				\toprule
				\multirow{1}{*}{\textbf{Methods}} &
				PSNR$\uparrow$ & SSIM$\uparrow$ & $D_{\mathrm{MAE}}\downarrow$ & $V_{\mathrm{RMSE}}\downarrow$ \\
				\midrule
				\multicolumn{3}{l}{\textbf{\textit{Recon. with labels}}} \\
				\textcolor{gray}{NeuRAD~\cite{tonderski2024neurad}}  & \textcolor{gray}{18.63} & \textcolor{gray}{0.458} & \textcolor{gray}{2.82} &	\textcolor{gray}{--} \\ %
				\textcolor{gray}{StreetGS~\cite{yan2024street}}  &\textcolor{gray}{16.35} &  \textcolor{gray}{0.398} & \textcolor{gray}{1.67} & \textcolor{gray}{--} \\ %
				\textcolor{gray}{G3R~\cite{chen2025g3r}} & \textcolor{gray}{18.93} &	\textcolor{gray}{0.501}	& \textcolor{gray}{2.41} & \textcolor{gray}{--}	 \\ %
				\midrule
				\multicolumn{3}{l}{\textbf{\textit{Unsupervised recon.}}}\\
				EmerNeRF$^\dagger$~\cite{yang2023emernerf} &  15.40 &	0.313	& 7.92 & 0.304  \\ %
				DeSiRe-GS$^\dagger$~\cite{peng2024desire} & 15.69 &  0.346 &  14.74 & 0.303 \\ %
				STORM$^\dagger$~\cite{yang2025storm} & 16.61	& 0.338	& 5.82 & 0.218 \\ %
				\textbf{\name-\textit{base} (Ours)} & 17.88 & 0.403 & 2.62 & 0.171 \\ %
				\textbf{\name (Ours)} & \textbf{19.07}  & \textbf{0.489} & \textbf{2.57} & \textbf{0.162} \\ %
				\bottomrule
			\end{tabular}
		}
	\end{minipage}
	\vspace{-0.1in}
\end{table}

\begin{figure*}[t]
	\centering
	\includegraphics[width=1.0\textwidth]{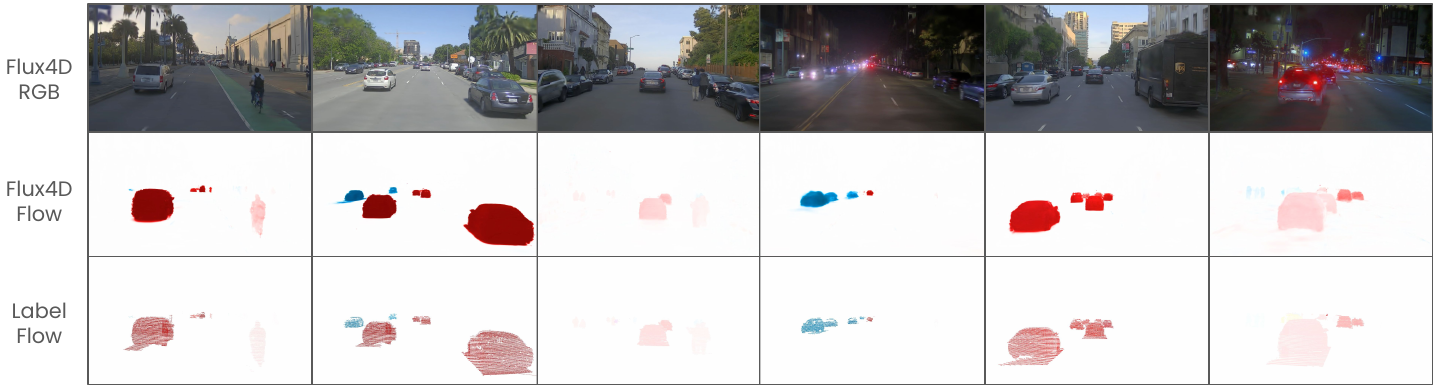}
	\vspace{-0.25in}
	\caption{\textbf{High-fidelity flow and RGB reconstruction.} \name not only provides photorealistic reconstruction of the dynamic scene but also estimates actors' motion flow with high precision.
	}
	\vspace{-0.05in}
	\label{fig:rgb_flow}
\end{figure*}

\paragraph{Novel view synthesis on PandaSet:}
Table~\ref{tab:comparison} and Fig.~\ref{fig:comp_1s} compare \name against SoTA unsupervised methods on 1s PandaSet snippets in the interpolation setting, with supervised approaches included for reference.
Reconstruction speed is measured on a single RTX A5000 GPU (24GB).
\name achieves superior photorealism and geometric accuracy with fast reconstruction speed.
We further evaluate our method on longer-horizon reconstruction of 8 second logs (Table~\ref{tab:comparison_full_snippet} and Fig.~\ref{fig:comp_8s}), using iterative processing of 1s snippets.
Our approach outperforms unsupervised per-scene optimization methods by a large margin on both 1s and 8s reconstruction tasks, without requiring pre-trained models or complex regularization. Our quantitative results as reported in these tables also indicate that \name is competitive even against supervised approaches.
Qualitatively, as shown in Fig.~\ref{fig:comp_1s} and~\ref{fig:comp_8s}, \name achieves high-fidelity camera rendering in both static and dynamic regions, while existing unsupervised approaches usually suffer from noticeable artifacts on dynamic actors due to inaccurate learned dynamics.

\vspace{-0.1in}
\paragraph{Novel view synthesis on WOD:}
We further compare \name with SoTA generalizable methods on WOD in Table~\ref{tab:drivingrecon}, where we {follow the setup} in~\cite{lu2024drivingrecon}. The baseline results are from DrivingRecon~\cite{lu2024drivingrecon} paper and we confirmed the setup and results with the authors to ensure accurate comparison. \name surpasses DrivingRecon by +5.99 dB in PSNR and +0.21 in SSIM, demonstrating its effectiveness for unsupervised dynamic scene reconstruction.
Please see Appendix~\ref{sec:additional_exp} for qualitative comparisons.

\vspace{-0.1in}
\paragraph{Flow estimation:}
We compare the estimated motion flows of \name with existing unsupervised per-scene optimization methods EmerNeRF~\cite{yang2023emernerf} and DeSiRe-GS~\cite{peng2024desire}. As shown in Table~\ref{tab:comparison},~\ref{tab:comparison_full_snippet} and Fig.~\ref{fig:comp_flow}, \name significantly outperforms prior approaches, learning accurate motion direction and magnitude without any supervision. In contrast, existing methods struggle to learn consistent motion flows and fully decompose dynamic scenes, leading to inaccurate and incoherent motion predictions, limiting their applicability in downstream tasks. 

\vspace{-0.1in}
\paragraph{Scene flow evaluation:}
While \name primarily focuses on reconstruction and is not specifically designed for scene flow estimation, we further evaluate its performance on PandaSet compared with representative scene flow estimation methods using standard scene flow metrics in Table~\ref{tab:scene_flow_pandaset} and~\ref{tab:scene_flow_pandaset2}. Please see Appendix~\ref{sec:additional_exp} for comparisons on WOD.
Although not designed for scene flow estimation, Flux4D achieves superior performance across most scene flow metrics using only reconstruction-based supervision (RGB + depth). Notably, it outperforms other methods on smaller or less common object categories such as wheeled VRUs, other vehicles, and pedestrians, as shown in bucketed evaluations.
These results highlight a promising path to unifying state-of-the-art scene flow estimation~\cite{khatri2024can,kim2025flow4d,li2025uniflow} and reconstruction within a single framework.

\vspace{-0.1in}
\paragraph{Future prediction:}
We evaluate \name's capability for future frame prediction beyond the observed frames.
This challenging task requires precise motion estimation, temporal consistency, occlusion reasoning, and a comprehensive 4D scene understanding.
As shown in Table~\ref{tab:future_prediction}, \name outperforms existing unsupervised methods in both photometric accuracy and geometric consistency. Moreover, \name even outperforms supervised approaches that rely on imperfect explicit annotations for extrapolation, demonstrating the robustness of our predicted scene representation and the effectiveness of unsupervised scene flow prediction.
This highlights \name{}'s ability to model scene dynamics, which is critical for world modeling, simulation, and scene understanding in autonomous systems.
We report dynamic-only metrics in Table~\ref{tab:future_prediction} and defer full-image metrics to Appendix~\ref{sec:additional_exp}.

\begin{table}[t]
	\centering
	\caption{\textbf{Comparison with scene flow estimation methods.} }
	\label{tab:scene_flow_pandaset}
	\vspace{-0.05in}
	\resizebox{\textwidth}{!}{
	\begin{tabular}{lccccccccc}
		\toprule
		\textbf{Method} & {EPE3D} $\downarrow$ & $Acc_{5} \uparrow$ & $Acc_{10} \uparrow$ & $\theta_\epsilon \downarrow$ & {EPE-BS} $\downarrow$ & {EPE-FS} $\downarrow$ & {EPE-FD} $\downarrow$ & {EPE-3way} $\downarrow$ & {Inference time} $\downarrow$ \\ \midrule
		NSFP~\cite{li2021neural}   & 0.183 & 0.558 & 0.713 & 0.510 & 0.106 & 0.103 & 0.573 & 0.227 & $\sim$5.57 s/frame \\
		FastNSF~\cite{li2023fast} & 0.194 & 0.571 & 0.714 & 0.471 & 0.155 & 0.134 & 0.428 & 0.211 & $\sim$0.68 s/frame \\
		STORM~\cite{yang2025storm}  & 0.120 & 0.757 & 0.782 & 0.489 & \textbf{0.009} & \textbf{0.098} & 0.536 & 0.201 & \textbf{$\sim$0.01 s/frame} \\
		\textit{Flux4D} & \textbf{0.094} & \textbf{0.775} & \textbf{0.807} & \textbf{0.123} & 0.019 & 0.117 & \textbf{0.391} & \textbf{0.165} & $\sim$0.31 s/frame \\ \bottomrule
	\end{tabular}
	}
\end{table}

\begin{table}[t]
	\centering
\caption{\textbf{Bucketed scene flow error on PandaSet.} Normalized EPE3D ($\downarrow$) per class, split into static (S) and dynamic (D) regions. Mean S/D are averages across all buckets. Abbrev.: BG = Background, CAR = Car, WVRU = Wheeled VRU, VEH = Other Vehicles, PED = Pedestrian.}
	\label{tab:scene_flow_pandaset2}
	\vspace{-0.05in}
	\resizebox{\textwidth}{!}{
	\begin{tabular}{lccccccccccc}
		\toprule
		\textbf{Method} & {BG-S}$\downarrow$ & {CAR-S}$\downarrow$ & {CAR-D}$\downarrow$ & {WVRU-S}$\downarrow$ & {WVRU-D}$\downarrow$ & {VEH-S}$\downarrow$ & {VEH-D}$\downarrow$ & {PED-S}$\downarrow$ & {PED-D}$\downarrow$ & {Mean S}$\downarrow$ & {Mean D}$\downarrow$ \\
		\midrule
		NSFP~\cite{li2021neural}    & 0.128 & 0.093 & 0.668 & 0.046 & 0.975 & 0.060 & 0.819 & 0.071 & 0.945 & 0.080 & 0.852 \\
		FastNSF~\cite{li2023fast} & 0.196 & 0.153 & \textbf{0.581} & 0.043 & 0.960 & 0.075 & 0.701 & 0.041 & \textbf{0.894} & 0.102 & \textbf{0.784} \\
		STORM~\cite{yang2025storm}   & \textbf{0.005} & 0.087 & 0.713 & \textbf{0.000} & 1.000 & 0.195 & 1.000 & 0.093 & 1.012 & 0.076 & 0.931 \\
		\name  & 0.019 & \textbf{0.078} & 0.701 & 0.011 & \textbf{0.866} & \textbf{0.021} & \textbf{0.661} & \textbf{0.027} & 0.966 & \textbf{0.031} & 0.800 \\
		\bottomrule
	\end{tabular}
}
\end{table}

\begin{table}[t]
	\centering
	\begin{minipage}{0.49\linewidth}
		\centering
		\caption{\textbf{Ablation study on \textit{Flux4D} designs.}}
		\label{tab:ablation}
		\vspace{-0.05in}
		\resizebox{\linewidth}{!}{
			\begin{tabular}{lcccc}
				\toprule
				\multirow{2}{*}{\textbf{Methods}} & \multicolumn{4}{c}{\textbf{Dynamic-only}} \\
				& PSNR$\uparrow$ & SSIM$\uparrow$ & $D_{\mathrm{MAE}}\downarrow$ & $V_{\mathrm{RMSE}}\downarrow$ \\
				\midrule
				\textit{\name-base} & 18.62 &	0.454 & 2.06 & 0.181 \\
				$+$ iterative refinement & 21.32 & 0.636 & 1.66 & \textbf{0.167} \\
				\ $+$ polynomial motion & \textbf{21.45} & \textbf{0.641} & \textbf{1.55} & \textbf{0.167} \\
				\bottomrule
			\end{tabular}
		}
	\end{minipage}%
	\hfill
	\begin{minipage}{0.49\linewidth}
		\centering
		\caption{\textbf{Ablation study on training strategy.}}
		\label{tab:ablation_loss}
		\vspace{-0.05in}
		\resizebox{\linewidth}{!}{
			\begin{tabular}{lcccc}
				\toprule
				\multirow{2}{*}{\textbf{Methods}} & \multicolumn{4}{c}{\textbf{Dynamic-only}} \\
				& PSNR$\uparrow$ & SSIM$\uparrow$ & $D_{\mathrm{MAE}}\downarrow$ & $V_{\mathrm{RMSE}}\downarrow$ \\
				\midrule
				\name & \textbf{21.99} & \textbf{0.662} & 1.63 & \textbf{0.157} \\
				$-$ vel. reweighting & 21.45 & 0.641 & 1.55 & 0.167 \\
				\ \ $-$ vel. regularization & 21.08 & 0.614 & \textbf{1.44} & 0.532 \\
				\bottomrule
			\end{tabular}
		}
	\end{minipage}
	\vspace{-0.1in}
\end{table}

\vspace{-0.1in}
\paragraph{Ablation:}
Table~\ref{tab:ablation} evaluates \name's key design components.
Iterative refinement and polynomial motion significantly enhance image quality, geometric accuracy and motion prediction performance.
Table~\ref{tab:ablation_loss} demonstrates that our static-preference prior is essential to learning accurate flow, and that velocity reweighting improves performance on the dynamic elements.
Please refer to Appendix~\ref{sec:additional_exp} for qualitative comparisons. 

\vspace{-0.1in}
\paragraph{LiDAR-free \name:}
We show that \name{} can also operate in a LiDAR-free mode at inference similar to DrivingRecon~\cite{lu2024drivingrecon} and STORM~\cite{yang2025storm} by using off-the-shelf monocular depth estimation model~\cite{hu2024metric3d}. As shown in Table~\ref{tab:monocular}, the flow estimation performance remains comparable, and in some cases, the visual realism improves in background regions (\textit{e.g.}, buildings) due to the broader coverage provided by monocular depth, particularly in areas where LiDAR sparsity limits reconstruction quality. Combining both LiDAR and points lifted by monocular depth yields the best overall realism.

\begin{table}[htbp!]
	\centering
	\caption{\textbf{LiDAR-free \name using off-the-shelf monocular depth estimation model~\cite{hu2024metric3d}}.}
	\label{tab:monocular}
	\vspace{-0.05in}
	\resizebox{\textwidth}{!}{
		\begin{tabular}{lccccccccccc}
			\toprule
			\multirow{2}{*}{\textbf{Methods}} & \multicolumn{4}{c}{\textbf{Dynamic-only}} & \multicolumn{4}{c}{\textbf{Full image}} & \textbf{Scene Flow} \\
			& PSNR $\uparrow$ & SSIM $\uparrow$ & $D_\mathrm{MAE}$\,$\downarrow$ & $V_\mathrm{RMSE}$\,$\downarrow$ & PSNR\,$\uparrow$ & SSIM\,$\uparrow$ & $D_\mathrm{MAE}$\,$\downarrow$ & $V_\mathrm{RMSE}$\,$\downarrow$ & EPE-3way\,$\downarrow$ \\
			\midrule
			\name (monocular depth only) & 21.71 & 0.668 & \textbf{1.45} & 0.159 & 23.87 & 0.688 & 1.23 & 0.186 & 0.165 \\
			\name (LiDAR, Table~\ref{tab:comparison}) & \textbf{21.99} & 0.662 & 1.63 & \textbf{0.157} & 23.84 & 0.675 & \textbf{1.07} & \textbf{0.182} & 0.165 \\
			\name (LiDAR + monocular depth) & \textbf{21.99} & \textbf{0.682} & 1.52  & 0.158 & \textbf{24.55} & \textbf{0.726} & 1.11 & 0.184 & \textbf{0.161} \\
			\bottomrule
	\end{tabular}}
\end{table}

\vspace{-0.1in}
\paragraph{Scaling analysis:}

\name's effectiveness stems from multi-scene training, leveraging diverse driving data as implicit regularization. Unlike per-scene methods that require complex regularizations or pre-trained models, increasing the amount of training data naturally improves scene decomposition and motion estimation. Analysis on PandaSet and WOD shows consistent improvements in photometric accuracy and motion estimation as training data scale. This confirms unsupervised 4D reconstruction benefits significantly from diverse real-world scenarios, suggesting \name can continue improving with additional data, making it promising for scalable scene reconstruction.

\begin{figure}[htbp!]
	\centering
	\includegraphics[width=1.0\textwidth]{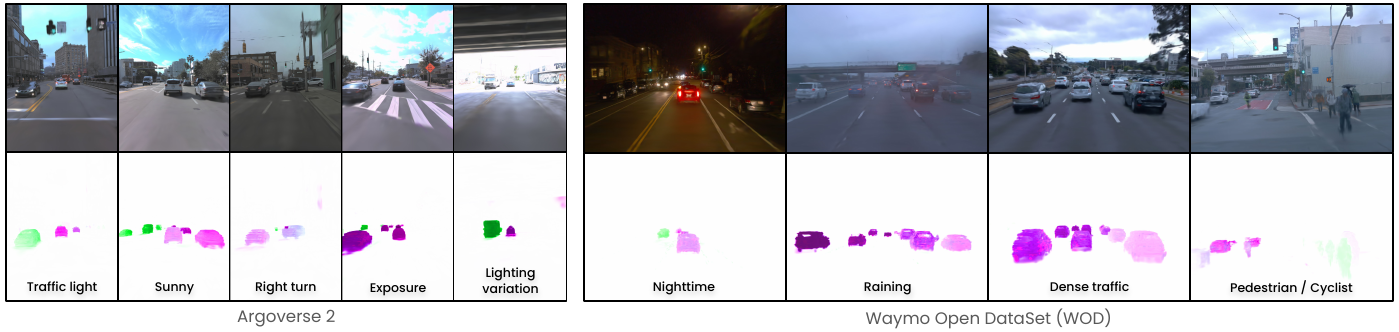}
	\vspace{-0.27in}
	\caption{\textbf{\name reconstruction on Argoverse 2 and WOD.}} %
\vspace{-0.12in}
\label{fig:argoverse_waymo}
\end{figure}

\begin{figure}[htbp!]
\centering
\begin{minipage}{0.45\linewidth}
	\includegraphics[width=\columnwidth]{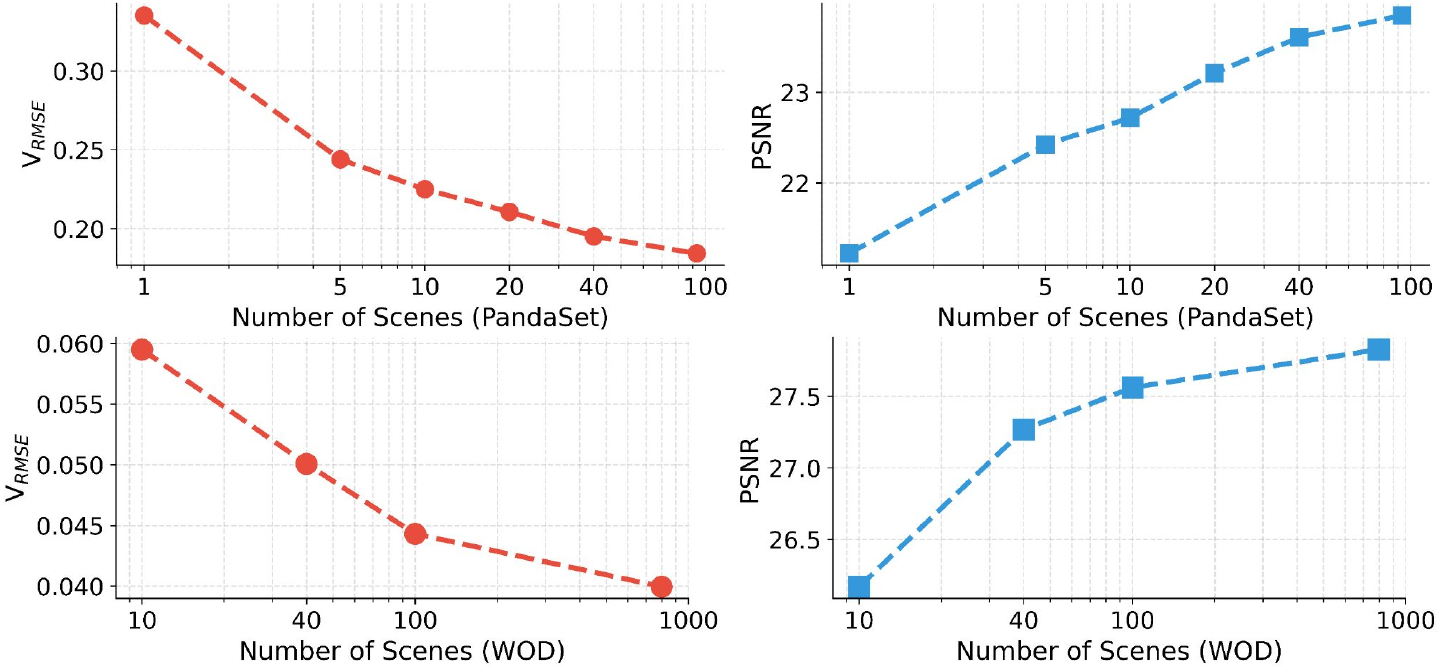}
	\vspace{-0.2in}
	\caption{\textbf{Scaling analysis.} Increasing number of training scenes for \textit{Flux4D} consistently improves performance.}
	\label{fig:scaling_laws}
\end{minipage}
\hfill
\begin{minipage}{0.53\linewidth}
	\includegraphics[width=\columnwidth]{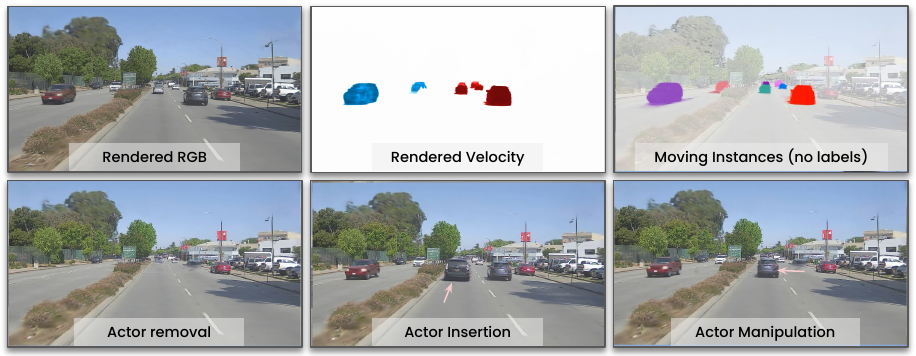}
	\vspace{-0.2in}
	\caption{\textbf{Simulation applications.} Flux4D can be applied suc-
		cessfully to different camera simulation tasks, e.g., actor removal,
		insertion and manipulation.}
	\label{fig:sim}
\end{minipage}
\end{figure}

\vspace{-0.1in}
\paragraph{Camera Simulation:}
We showcase applying \name for high-fidelity camera simulation in large-scale driving scenarios. \name produces high-quality motion flows in diverse, large-scale dynamic scenes on PandaSet (Fig.~\ref{fig:rgb_flow}), Argoverse 2 \cite{wilson2023argoverse}, and WOD (Fig.~\ref{fig:argoverse_waymo}). This allows accurate scene decomposition across diverse environments which is critical for instance extraction and direct manipulation of dynamic elements (Fig.~\ref{fig:sim}).
Compared to existing self-supervised per-scene methods, \name is better suited for interactive and controllable applications, as it reconstructs an editable representation that supports instance mask extraction, scene editing and object manipulation for various downstream tasks.
In Fig.~\ref{fig:sim}, we demonstrate \name's capability to render realistic images of the modified scene representation.
Notably, our approach achieves this without requiring labels. %

\section{Limitations}
\label{sec:limitations}

Although \name{} achieves SoTA 4D reconstruction without any annotations or pre-trained models, three key limitations remain: (1) flow estimation for highly dynamic actors with complex motion patterns is challenging, which could be mitigated by leveraging larger and more diverse training data; (2) iterative approach for long-horizon reconstruction creates visible inconsistencies at transition points; and (3) the method assumes a simple pinhole camera model with clean LiDAR data, limiting applicability with rolling shutter cameras or noisy sensor inputs. Please see Appendix~\ref{sec:supp_limitations} for more examples.
Future work will focus on scaling to larger datasets, developing a unified temporal representation for seamless long-term reconstruction, and improving robustness to real-world sensor imperfections.
Furthermore, \textit{Flux4D}'s explicit 3D representation offers interpretable structure for world models.
Overall, we believe that our simple and scalable design serves as a foundation for the community to build upon, enabling further advancements in 4D reconstruction.

\section{Conclusion}
\label{sec:conclusion}
We present \name, a scalable flow-based unsupervised framework for reconstructing large-scale dynamic scenes by directly predicting 3D Gaussians and their motion dynamics. By relying solely on photometric losses and enforcing an ``as static as possible'' regularization, \name effectively decomposes dynamic elements without requiring any supervision, pre-trained models, or foundational priors. Our method enables fast reconstruction, scales efficiently to large datasets, and generalizes well to unseen environments. Extensive experiments on outdoor driving datasets demonstrate state-of-the-art performance in scalability, generalization, and reconstruction quality. We hope this work paves the way for efficient, unsupervised 4D scene reconstruction at scale.

\section*{Acknowledgement}
We sincerely thank the anonymous reviewers for their insightful suggestions especially on scene flow evaluation, paper framing, and additional experiments using monocular depth estimation models. We would like to thank
Andrei Bârsan and Joyce Yang for their feedback on the early draft. We also thank the Waabi team
for their valuable assistance and support.

\bibliographystyle{plain}
\bibliography{main}

\newpage
\section*{Appendix}

\appendix
\renewcommand{\thetable}{A\arabic{table}}
\renewcommand{\thefigure}{A\arabic{figure}}

In this appendix, we provide detailed information on our method and the baselines, additional experimental results, and a discussion of limitations and future work.
	We begin by describing the implementation details of \name in Sec.~\ref{sec:additional_details}.
	After that, in Sec.~\ref{sec:baseline_details} we present how the baseline models were adapted to our experimental settings, which are described in Sec.~\ref{sec:exp_details}.
	Next, we show additional experimental results in Appendix~\ref{sec:additional_exp}, including future prediction with \name, ablation studies, and more results for controllable simulation.
	We also include additional discussions and comparisons of \name with closely related works in Sec.~\ref{sec:discussion}.
	Moreover, we provide the limitations of \name and future directions in Sec.~\ref{sec:supp_limitations}.
	Finally, we discuss \name's broader impact (Sec.~\ref{sec:broader_impact}), computation resources (Sec.~\ref{sec:computation}), and asset's licenses (Sec.~\ref{sec:license}).

\section{\name Details}
\label{sec:additional_details}

\subsection{Motion Enhancement Details}
\label{sec:motion_details}
\paragraph{Velocity reweighting:} To further enhance the flow estimation for dynamic actors, we employ a pixel-wise reweighting scheme in loss computation based on the predicted velocity magnitude. This is 
used in place of $\mathcal{L}_\text{rgb}$ during training, and is defined as:
\begin{align}
	\mathcal{L}_\text{rgb\_vw} &= (1 + ||sg(\mathbf{v_r})||) \cdot \mathcal{L}_\text{rgb},
\end{align}
where $sg(\cdot)$ denotes the stop gradient operation, $\mathbf{v_r}$ is the rendered velocity in image space, and $\mathcal{L}_\text{rgb}$ is the photometric loss.
For stability, we clip the values of $\mathbf{v_r}$ to $[0, 10]$. 

\vspace{-0.15in}
\paragraph{Polynomial motion modeling:} In addition to the constant velocity assumption, we also consider a polynomial motion model to better handle more complex actor behaviors. The velocity is parameterized as a 
polynomial function of degree $\ell$. Specifically, rather than predicting a single constant velocity, we predict polynomial velocity parameters $\mathbf v = [\mathbf v_0, \mathbf v_1, \ldots, \mathbf v_\ell]$, such that 
\begin{align}
	\mathbf p_i^{t'} = \mathbf p_i^{t_i} + \sum_{j=0}^{\ell} \mathbf v_j \frac{({t'}^{j+1} - t_{i}^{j+1})}{j+1}.
	\label{eq:movement_poly}
\end{align}
In our experiments, we set $\ell=1$, which corresponds to the constant acceleration assumption. We also explored higher-order terms with $\ell=2, 3$, but did not observe noticeable gains.

\subsection{Model Architecture}
\label{sec:arch}
\paragraph{Initialization:} 
Given a sequence of input images and corresponding LiDAR point clouds, we initialize the Gaussian parameters as follows:

\begin{itemize}[left=5pt]
\item \vspace{-0.25em} \textbf{Position:} Each Gaussian is initialized at the location of its corresponding LiDAR point.
\item \vspace{-0.25em} \textbf{Color:} For each image at a given frame, we project the LiDAR points at that frame onto the image and sample their colors from the image.
\item \vspace{-0.25em} \textbf{Scale:} The scale of each Gaussian is set to the average distance to its three nearest neighbors. Following \cite{chen2024omnire,wei2024meshlrm}, we apply a logarithmic transformation and \texttt{softplus} activation to stabilize training.
\item \vspace{-0.25em} \textbf{Orientation:} The orientation of each Gaussian  is randomly initialized and then normalized.
\item \vspace{-0.25em} \textbf{Opacity:} The opacity of each Gaussian is initialized to 0.5.
\item \vspace{-0.25em} \textbf{Time:} Each Gaussian inherits the timestamp of its corresponding LiDAR point, which is then normalized to the range $[0, 1]$.
\end{itemize}

\begin{figure*}[t]
	\centering
	\includegraphics[width=1.0\textwidth]{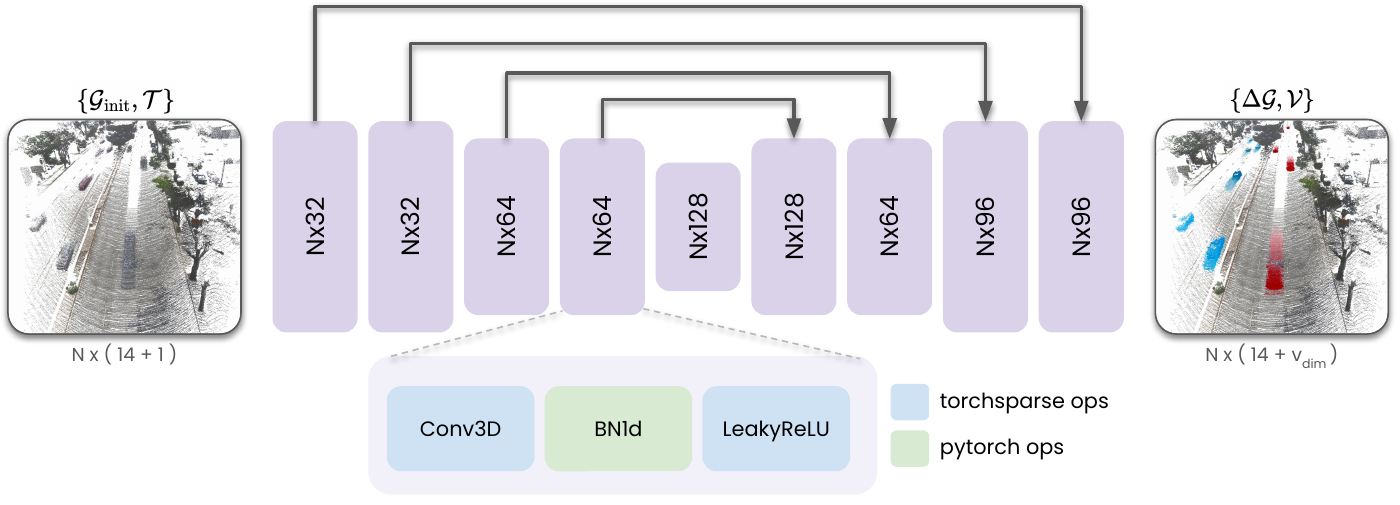}
	\vspace{-0.25in}
	\caption{\textbf{\textit{Flux4D-base} model architecture.}
	}
	\label{fig:flux4dbase_arch}
\end{figure*}

To model the sky, we compute the axis-aligned bounding box (AABB) of the initialized Gaussians above, and randomly sample 1 million sky points in the upper-half of the sphere centered at the center of the AABB, with radius set to 4 times the length of the AABB. %

\paragraph{Flux4D-base:} We employ a Sparse 3D UNet (Fig. \ref{fig:flux4dbase_arch}) adopted from \cite{tangandyang2023torchsparse} for our Flux4D-base reconstruction model $f_\theta$. The input to $f_\theta$ are $\mathcal{G}_{\mathrm{init}} = \{  \mathbf{g}_{\mathrm{init}, i} :  \mathbf{g}_{\mathrm{init}, i} \in \mathbb{R}^{14}\}$ and $\mathcal{T} = \{{t}_i: {t}_i \in \mathbb{R}^{1}\}$. The first 3 channels of each $\mathbf{g}_i$ are the 3D Gaussian location ($\mathbf{p}_i\in\mathbb{R}^3$), following by 4 channels for quaternions ($\mathbf{q}_i\in\mathbb{R}^4$), 3 channels for scales ($\mathbf{s}_i\in\mathbb{R}^3$), 1 channel for opacity ($\mathbf{o}_i\in\mathbb{R}^1$), and the last 3 for colors ($\mathbf{c}_i\in\mathbb{R}^3$). All input are concatenate on the last dimension, resulting in the input of tensor of shape $N\times14$. Although we observe a marginal improvement in reconstruction quality with spherical harmonics, we omit them from our representation for simplicity, with no compromise in motion quality.
Each block contains a \texttt{spnn.Conv3D} layer follows by \texttt{nn.BatchNorm1d} and \texttt{spnn.LeakyReLU}, where \texttt{nn} is the standard \texttt{PyTorch} \cite{paszke2019pytorch} operation and \texttt{spnn} is the \texttt{Torchsparse} \cite{tangandyang2023torchsparse} operation. The skip connections are applied by concatenating down features with up features channel-wise. 
The output are the per-gaussian residues $\Delta \mathcal{G} = \{\Delta \mathbf{g}_i: \Delta \mathbf{g}_i \in \mathbb{R}^{14}\}$ and velocity $\mathcal{V} = \{\mathbf{v}_i: \mathbf{v}_i \in \mathbb{R}^{v_\text{dim}}\}$. 
Here, $v_\text{dim} = 3(\ell + 1)$ following the convention from Sec. \ref{sec:motion_details}, where $\ell=0$ for constant velocity assumption and $\ell=1$ for constant acceleration.  

\paragraph{Rendering:} To render the scene at a given time $t$, for each inital Gaussian $\mathbf{g}_{\mathrm{init}, i}$, we first apply the residue $\Delta \mathbf{g}_i$ to obtained the refined gaussian $\mathbf{g}_i$:
\begin{align*}
	\mathbf{g}_i &= \mathbf{g}_{\mathrm{init}, i} + \Delta \mathbf{g}_i = \{\textbf{p}_{\mathrm{init}, i} + \Delta \textbf{p}_i, \textbf{q}_{\mathrm{init}, i} + \Delta \textbf{q}_i, \textbf{s}_{\mathrm{init}, i} + \Delta \textbf{s}_i, \textbf{o}_{\mathrm{init}, i} + \Delta \textbf{o}_i, \textbf{c}_{\mathrm{init}, i} + \Delta \textbf{c}_i,\},
\end{align*}
Afterwards, we apply the velocity $\mathbf{v}_i$ to the Gaussian position $p_i$ to obtain the position at time $t$ following Eqn.~(\ref{eq:movement_poly}), obtaining the final $\mathcal{G}_t$. We use \texttt{gsplat} \cite{gsplat} for Gaussian rendering. Before passing $\mathcal{G}_t$ to \texttt{gsplat}, we activate the Gausisan parameters by appling sigmoid function to the opacity and color of each Gaussian. We also clamp the scale to $[0m, 1m]$ to stabilize the training.

\paragraph{Iterative refinement:} 
For iterative refinement module $r_\theta$, we use the same UNet architecture as above, but take the gradient of the previous step's residual as additional input. We denote $\nabla \mathcal{G}^{(i)}$ as the gradient of the Gaussian parameters after $i^{th}$ iteration ($\mathcal{G}^{(i)}$), and is obtained by storing the detached gradients of $\Delta \mathcal{G}^{(i)}$ after the backward pass. Here, $\Delta \mathcal{G}^{(i)}$ is the output residue of $r_\theta$ after the $i^{th}$ iteration. For the \textit{Flux4D} model with $N$ iteration, the final refined Gaussian $\mathcal{G}$ is obtained by:
\begin{align*}
	\mathcal{G} &= \mathcal{G}_\mathrm{init} +  \sum_{j=1}^{N - 1} \Delta \mathcal{G}_i^{(j)} =  \{\mathbf{g}_i: \mathbf{g}_i = \mathbf{g}_{\mathrm{init}, i} + \sum_{j=1}^{N - 1} \Delta \mathbf{g}_i^{(j)}\},
\end{align*}
where
\begin{align*}
	\Delta 	\mathcal{G}_i^{(j)} = 
	\begin{cases}
		f_\theta(\mathcal{G}_{\mathrm{init}}, \mathcal{T}), & \text{if } j = 0, \\
		r_\theta(\mathcal{G}^{(j-1)}, \mathcal{T}, \nabla \mathcal{G}^{(j-1)}), & \text{if } j > 0.
	\end{cases}
\end{align*}
In all experiments, we set $N=3$. We observe that increasing $N$ (\textit{e.g.}, $N = 10$) can improve reconstruction quality but comes at the cost of reduced inference speed.

\subsection{Pseudocode}
\paragraph{\name:} We provide the pytorch-style pseudocode of \name in Algorithm~\ref{alg:code}. 

\begin{algorithm}[]
	\caption{\textbf{Pseudocode of \name in PyTorch style.}}
	\label{alg:code}
	\begin{lstlisting}[language=Python]
# Network: neural network for predicting Gaussians
# G_init: Gaussians initialized from LiDAR+image, [N x 14]
# T: per-Gaussian normalized time captured, [N x 1]
# G_delta: predicted Gaussians residue, [N x 14]
# V: velocity of each Gaussian, [N x v_dim]

# Predict updated Gaussians
G_delta, V = Network(G_init, T)

# Apply residue to Gaussians
G = G_init + G_delta 

L_recon = 0
for t_target in range(F): # Iterate over each frame
    # Move Gaussians to time t using velocities
    G_t = G.clone()
    G_t.means = G.means + V * (t_target - G.t)

    # Render scene with updated Gaussians
    I_rendered_t = render_gaussians(G_t) # Nx(3+1)

    # Compute losses
    L_recon += F.l1_loss(I_rendered_t, I_gt_t)

# velocity regularization
L_vel = torch.mean(torch.norm(V, p=2, dim=-1))

# Backpropagation and optimization
loss = L_vel + L_recon
optimizer.zero_grad()
loss.backward()
optimizer.step()
	\end{lstlisting}
\end{algorithm}

\paragraph{Iterative refinement:}
The pseudocode for iterative refinement as described in Sec.~\ref{sec:arch} can be found in Algorithm~\ref{alg:iter_code}.

\begin{algorithm}[]
	\caption{\textbf{Pseudocode for Iterative Refinement in Pytorch style.}}
	\label{alg:iter_code}
	\begin{lstlisting}[language=Python]
# F_Network: Flux4D-base neural network (f_theta)
# R_Network: FLux4D refinement network (r_theta)
# N: number of iterations, we fix N = 3

G = G_init
G_grad, V = None, None

# Iterative refinement
for step in range(N):
    # Predict first iteration's residue
    if step == 0:
        G_delta, V = F_Network(G, T)

    # Predict subsequent iterations' residue
    else:
        G_delta = R_Network(G, T, G_grad)

    # Apply residue to Gaussians
    G = G + G_delta

    # Compute Loss following Flux4D Pseudocode
    L = compute_loss(G, T, V)

    # Backpropagation
    L.backward()

    # Store gradients for next iteration
    G_grad = G.grad.clone()

	\end{lstlisting}
\end{algorithm}

\paragraph{Camera simulation:}
Additionally, we provide the pseudocode for dynamic clustering as seen in Fig.~\ref{fig:supp_sim},~\ref{fig:supp_sim2} in Algorithm~\ref{alg:sim_code}.

\begin{algorithm}[]
	\caption{\textbf{Pseudocode for Dynamic Clustering in Python style.}}
	\label{alg:sim_code}
	\begin{lstlisting}[language=Python]
# G: the refined Gaussian set (i.e., after applying the residue), [N x 14]
# V: per-Gaussian velocity, [N x 3]
# V_thresh: threshold for dynamic filtering, we fix V_thresh=5

# Compute the velocity norm of each Gaussian
V_normed = torch.norm(V, dim=1)

# Obtain dynamic Gaussians
dynamic_G = G[V_normed > V_thresh]

# Perform clustering on dynamic_G's positions obtain moving instances
instances = DBSCAN(eps=0.4, min_samples=10).fit_predict(G.means) # NxC
unique_instances = torch.unique(instances)

# Obtain per-instance Gaussians
instance_G = []
for instance_id in unique_instances:
    instance_G.append(G[instances == instance_id])

	\end{lstlisting}
\end{algorithm}

\section{Baseline Implementation Details}
\label{sec:baseline_details}

\paragraph{NeuRAD~\cite{tonderski2024neurad}:}
NeuRAD models dynamic scenes using compositional neural radiance fields while introducing techniques to handle complex sensor effects such as rolling shutter, ray dropping, and beam divergence. We use the publicly available implementation\footnote{\url{https://github.com/georghess/neurad-studio}} in \texttt{neurad} mode and train the models for 10,000 iterations to ensure convergence in 1.5s snippets. Training for only 5,000 iterations results in noticeable color artifacts, likely due to insufficient adaptation of the convolutional neural network. For 8s full-log reconstruction, we train for 20,000 iterations.

\vspace{-0.1in}
\paragraph{Street Gaussian~\cite{yan2024street}:}
Street Gaussian replaces NeRF-based representations~\cite{unisim,tonderski2024neurad} with compositional 3D Gaussian Splatting (3DGS), enabling real-time camera simulation. We adopt \texttt{drivestudio}'s offical implementation\footnote{\url{https://github.com/ziyc/drivestudio}} with default hyperparameters, including density control and the learning rate schedule. The 3D Gaussians are initialized with 800,000 downsampled aggregated LiDAR points and a randomly sampled subset of 200,000 points. The model is trained for 10,000 iterations for 1.5s snippet reconstruction and 20,000 iterations for 8s full-log reconstruction.

\vspace{-0.1in}
\paragraph{EmerNeRF~\cite{yang2023emernerf}:}
EmerNeRF is a self-supervised method for reconstructing 4D neural scene representations. It decomposes static and dynamic components while learning 3D scene flows without relying on ground truth object annotations. We adopt the public implementation\footnote{\url{https://github.com/NVlabs/EmerNeRF}} and follow the configuration incorporating a dynamic encoder and flow encoder. %
Following OmniRe~\cite{chen2024omnire}, we use SegFormer~\cite{xie2021segformer} to extract sky masks for training. The model is trained for 10,000 iterations for 1.5s snippet reconstruction and 20,000 iterations for 8s full-log reconstruction.

\vspace{-0.1in}
\paragraph{DeSiRe-GS~\cite{peng2024desire}:}
DeSiRe-GS is a 3DGS-based representation designed for self-supervised static-dynamic decomposition and high-quality surface reconstruction in driving scenes. It follows a two-stage pipeline: the first stage extracts 2D motion masks by computing feature differences between rendered and ground-truth images, while the second stage distills this 2D motion information into Gaussian space using PVG~\cite{chen2023periodic}. We use the public implementation\footnote{\url{https://github.com/chengweialan/desire-gs}} and train for 5,000 iterations per stage for 1.5s snippet reconstruction. For 8s full-log reconstruction, we train for 20,000 iterations in the first stage and 30,000 iterations in the second stage.

\vspace{-0.1in}
\paragraph{L4GM~\cite{ren2025l4gm}:} 
We fine-tune the officially released L4GM \texttt{big} model on PandaSet~\cite{xiao2021pandaset} for novel view synthesis (interpolation). Each training sample consists of 11 consecutive frames with rescaled cameras and scene depth, ensuring that the furthest camera from the scene center matches L4GM's default camera radius of 1.5. Our modified pipeline uses a single camera view per frame, taking six even-numbered frames as input to predict all frames in the sequence. The network is supervised with photometric and LPIPS losses, along with a 0.1-weighted depth loss aligning ray termination with ground truth LiDAR depth. While depth supervision slightly improves reconstruction quality, higher weights lead to catastrophic forgetting. Despite this, the predicted depth remains low-range, mostly capturing forward-facing scenes with flat geometry.

\vspace{-0.1in}
\paragraph{DepthSplat~\cite{xu2024depthsplat}:} 
We implement a feed-forward, multi-view generalizable 3DGS baseline inspired by~\cite{xu2024depthsplat}. Our model employs a U-Net with residual blocks to process six input frames, each containing RGB, depth, and 3D point information. To enhance geometric accuracy in outdoor scenes, we incorporate LiDAR data as a depth prior by aggregating LiDAR points across multiple frames and rendering them as depth maps for additional input features. The network predicts Gaussian parameters per pixel. For dynamic scenes, we extend the model with motion modeling by predicting velocity. The input is augmented with a binary dynamic mask to identify moving objects, while the output is expanded to a velocity-augmented Gaussian representation, including a 3D velocity vector. During rendering, point positions are updated similarly to \name{}. We constrain velocities using a \texttt{tanh} activation and introduce the same ``as static as possible'' regularization for velocity.
 
\vspace{-0.1in}
\paragraph{G3R~\cite{chen2025g3r}:}
Gradient guided generalizable reconstruction (G3R) is a framework that enables efficient and high-quality 3D scene reconstruction for large-scale scenes using iterative gradient feedback.
In our implementation of G3R, we adhere to a source-novel split: six frames serve as the source to compute gradient input, four frames for interpolation, and five frames for extrapolation during supervised training. We utilize SparseUNet from \texttt{torchsparse}, training the model for 24,000 ($1000 \times 24$) iterations. 
Gaussians are initialized by aggregating LiDAR points, utilizing labels for both static backgrounds and dynamic actors. We use labels to transform actor gaussians during rendering. 
For full 8-second reconstructions, we follow the same setting as~\cite{chen2025g3r}. For 1-second reconstructions, we limit the number of points to 800,000 to better reflect real-world conditions.

\vspace{-0.1in}
\paragraph{STORM~\cite{yang2025storm}:}
STORM is a generalizable, self-supervised method for dynamic scene reconstruction from multi-view RGB. It employs a feed-forward transformer to jointly predict per-pixel 3D Gaussian primitives and scene flow. We use the authors' public implementation\footnote{\url{https://github.com/NVlabs/GaussianSTORM}} with default hyperparameters unless noted, and fine-tune for 100{,}000 iterations to reconstruct 1.5\,s snippets on 8$\times$H100 GPUs with a global batch size of 32 (image resolution: $270\times480$). We also fine-tune the released pre-trained checkpoint on PandaSet. Both training regimes yield comparable performance.

\section{Experiment Details}
\label{sec:exp_details}
\subsection{Dataset Splits}

\paragraph{Pandaset:} For all experiments with Pandaset \cite{xiao2021pandaset}, namely interpolation (Table~\ref{tab:comparison},~\ref{tab:scene_flow_pandaset},~\ref{tab:scene_flow_pandaset2},~\ref{tab:monocular}), full-log reconstruction (Table~\ref{tab:comparison_full_snippet}), future prediction (Table~\ref{tab:future_prediction}), scaling law (Fig.~\ref{fig:scaling_laws} top), and ablation (Table~\ref{tab:ablation},~\ref{tab:ablation_loss}), we use logs $\texttt{001,011,016,065,084,090,106,115,123,158}$ for testing and the remaining for training following \cite{ze2023unisim,tonderski2024neurad}.

\paragraph{Waymo Open Datset:} We conduct experiment with DrivingRecon \cite{lu2024drivingrecon} on WOD \cite{waymo} in Table~\ref{tab:drivingrecon}. We adopt their data splits, combining the NOTR \textit{dynamic32} split and \textit{static32} split. These splits are curated by \cite{yang2023emernerf} and are publicly available at \url{https://github.com/NVlabs/EmerNeRF/blob/main/docs/NOTR.md}. 

\paragraph{Argoverse2:} We use Argoverse2 Sensor Dataset \cite{chang2019argoverse} to showcase our ability to handle multiple datasets in Fig.~\ref{fig:argoverse_waymo}. We process the Argoverse data using Omnire's open-source script at \url{https://github.com/ziyc/drivestudio/blob/main/datasets/argoverse/argoverse_preprocess.py}, and select the first 100 logs for testing and the rest for training. 

\subsection{Metrics}
\paragraph{Depth MAE:}
We evaluate geometric accuracy using Mean Absolute Error (MAE) against LiDAR-projected ground-truth depth maps.
For each pixel $i$, let $\hat{d}_i$ and $d_i$ denote the predicted and ground-truth depth, respectively.
We define the set of valid pixels as $\mathcal{D} = \{i \mid d_i > 0.01 \,\text{m} \;\wedge\; d_i < 80\,\text{m}\}$, filtering by ground-truth LiDAR hits only.
The depth MAE is then:
\begin{equation}
	{D}_\mathrm{MAE} = \frac{1}{|\mathcal{D}|} \sum_{i \in \mathcal{D}} \left| \hat{d}_i - d_i \right|.
\end{equation}
We choose MAE over Root Mean Square Error (RMSE) to reduce sensitivity to outliers caused by erroneous predictions on far-away background regions.

\paragraph{Normalized velocity RMSE:}
We measure scene flow accuracy using a spatio-temporally normalized RMSE over 3D velocity vectors.
Let $\hat{\mathbf{v}}_i, \mathbf{v}_i \in \mathbb{R}^3$ be the predicted and ground-truth 3D velocity at pixel $i$.
We define a binary validity mask $m_i = \mathds{1}[d_i > 0] \cdot r_i$, where $\mathds{1} [d_i > 0]$ indicates LiDAR hit pixels and $r_i$ is the region mask (dynamic or non-sky).
For each snippet with $T$ evaluation frames of resolution $H \times W$, the normalized velocity RMSE is:
\begin{equation}
	{V}_\mathrm{RMSE} = \sqrt{\frac{1}{T \cdot H \cdot W \cdot 3} \sum_{t=1}^{T}\sum_{j=1}^{H \times W} m_j^{(t)} \left\| \hat{\mathbf{v}}_j^{(t)} - \mathbf{v}_j^{(t)} \right\|^2},
\end{equation}
and the final score is averaged over all snippets.
The denominator includes all pixels and all timesteps rather than only valid ones: masked-out pixels contribute zero error but still count toward normalization.
This design makes the metric agnostic to both image resolution and sequence length, and naturally assigns greater weight to scenes with more dynamic content.

\subsection{Experiment Settings Details}
\paragraph{Interpolation:} 
We split the training sequences into 1-second snippets (11 frames), ensuring a 5-frame overlap between consecutive snippets. 
Frames at indices [0, 2, 4, 6, 8, 10] are used as input frames, while frames at indices [1, 3, 5, 7, 9] serve as target frames.

\vspace{-0.1in}
\paragraph{Future prediction:} We split the training sequences into 1.5-second snippets (16 frames), ensuring a 5-frame overlap between consecutive snippets. 
Frames at indices [0, 2, 4, 6, 8, 10] are used as input frames, while frames at indices [1, 3, 5, 7, 9, 11, 12, 13, 14, 15] serve as target frames.

\subsection{Training Details} We train both models for 30,000 steps for approximately 2 days on $4\times$ NVIDIA L40S GPUs. We use batch size 1 per GPU, 
achieving an effective batch size of 4. We use the Adam optimizer with a learning rate of 1e-3, with an exponential-decay annealing scheduler and a warmup phase of 1000 steps.
We use the full-resolution camera images on all datasets:  PandaSet ($1920 \times 1080$), Waymo Open Dataset ($1920 \times 1280$), Argoverse 2 ($1550 \times 2048$). 
For the scaling analysis, all models are trained on PandaSet for 10,000 steps and on WOD for 30,000 steps.

\section{Additional Experiments}
\label{sec:additional_exp}

\subsection{Scene Flow Evaluation}
Table~\ref{tab:wod_overall} and~\ref{tab:wod_buckets} evaluate the scene flow performance of FastNSF~\cite{li2023fast} and \name{} on WOD.
Although not designed for scene flow estimation, \name{} achieves superior performance across all scene flow metrics. We leave comparisons to state-of-the-art dedicated scene flow methods (\textit{e.g.}, EulerFlow~\cite{vedder2025neural}) in the future work. These findings unveil a promising path to unifying state-of-the-art flow and reconstruction within a single framework.

\begin{table}[htbp!]
	\centering
	\caption{\textbf{Comparison with scene flow estimation methods on WOD.}}
	\label{tab:wod_overall}
	\vspace{-0.05in}
	\resizebox{\textwidth}{!}{
		\begin{tabular}{lccccccccc}
			\toprule
			\textbf{Method} & {EPE3D} $\downarrow$ & $Acc_{5} \uparrow$ & $Acc_{10} \uparrow$ & $\theta_\epsilon \downarrow$ & {EPE\mbox{-}BS} $\downarrow$ & {EPE\mbox{-}FS} $\downarrow$ & {EPE\mbox{-}FD} $\downarrow$ & {EPE\mbox{-}3way} $\downarrow$ & {Inference time} $\downarrow$ \\
			\midrule
			FastNSF~\cite{li2023fast} & 0.162 & 0.734 & 0.805 & 0.908 & 0.131 & 0.076 & 0.650 & 0.203 & $\sim$0.28 s/frame \\
			\textit{Flux4D} & \textbf{0.048} & \textbf{0.901} & \textbf{0.929} & \textbf{0.540} & \textbf{0.011} & \textbf{0.012} & \textbf{0.440} & \textbf{0.114} & \textbf{$\sim$0.20 s/frame} \\
			\bottomrule
		\end{tabular}
	}
\end{table}

\begin{table}[htbp!]
	\centering
	\caption{\textbf{Bucketed scene flow error on WOD.} Normalized EPE3D ($\downarrow$) per class, split into static (S) and dynamic (D) regions. Mean S/D are averages across all buckets. Abbrev.: BG = Background, VEH = Vehicle, PED = Pedestrian, CYC = Cyclist.}
	\label{tab:wod_buckets}
	\vspace{-0.05in}
	\resizebox{\textwidth}{!}{
		\begin{tabular}{lccccccccc}
			\toprule
			\textbf{Method} & {BG\mbox{-}S}$\downarrow$ & {VEH\mbox{-}S}$\downarrow$ & {VEH\mbox{-}D}$\downarrow$ & {PED\mbox{-}S}$\downarrow$ & {PED\mbox{-}D}$\downarrow$ & {CYC\mbox{-}S}$\downarrow$ & {CYC\mbox{-}D}$\downarrow$ & {Mean S}$\downarrow$ & {Mean D}$\downarrow$ \\
			\midrule
			FastNSF~\cite{li2023fast} & 0.144 & 0.043 & 0.653 & 0.044 & 1.026 & \textbf{0.026} & 0.744 & 0.064 & 0.807 \\
			\textit{Flux4D} & \textbf{0.011} & \textbf{0.009} & \textbf{0.502} & \textbf{0.014} & \textbf{0.680} & \textbf{0.026} & \textbf{0.732} & \textbf{0.015} & \textbf{0.638} \\
			\bottomrule
		\end{tabular}
	}
\end{table}

\subsection{Ablation Study}
\begin{figure*}[t]
	\centering
	\includegraphics[width=1.0\textwidth]{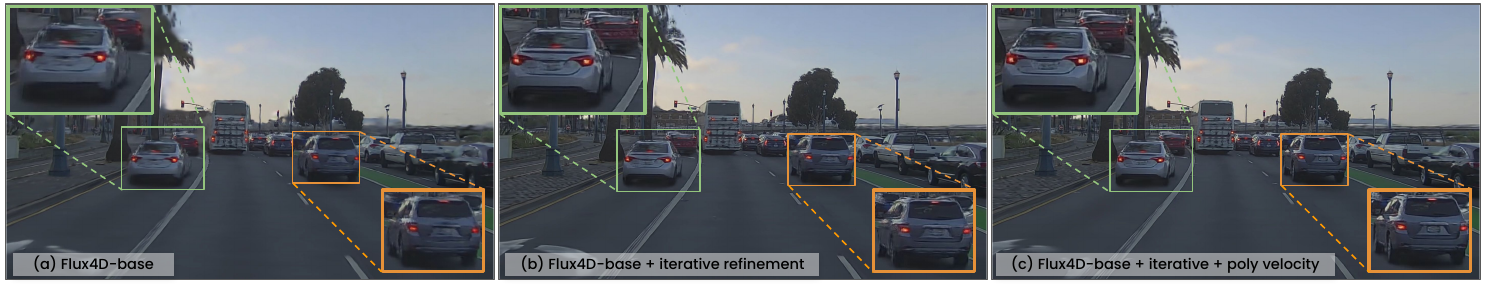}
	\vspace{-0.2in}
	\caption{\textbf{Ablation study on \name components.} Iterative refinement (2 iterations) and polynomial motion modeling (constant acceleration, $\ell=1$) enhance fine-grained reconstruction details.}
	\label{fig:supp_ablation1}
\end{figure*}

\paragraph{Ablation on \name components:} We conduct an ablation study on the key components of \name in Fig.~\ref{fig:supp_ablation1}.
The results clearly show that both iterative refinement and polynomial motion modeling enhance the sharpness and overall rendering quality. 

\begin{figure*}[t]
	\centering
	\includegraphics[width=1.0\textwidth]{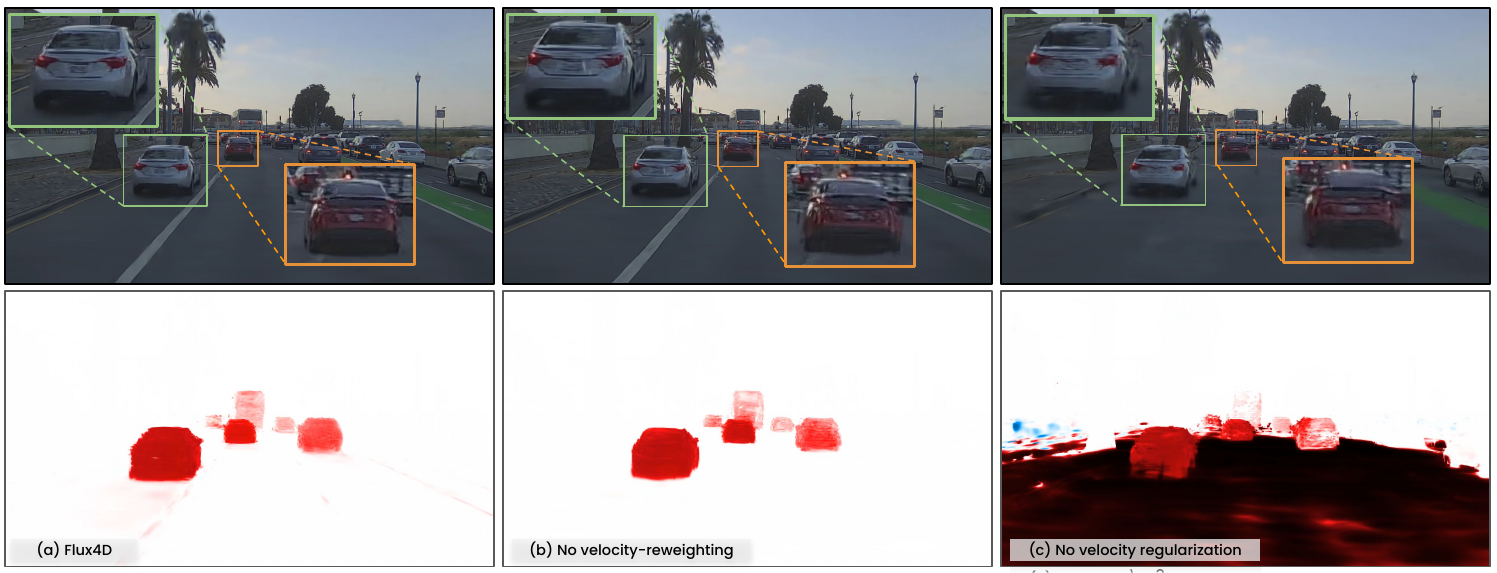}
	\vspace{-0.2in}
\caption{\textbf{Ablation study on training strategies.} Velocity re-weighting and static-preference regularization are crucial for capturing fine-grained details in dynamic actors and ensuring accurate, consistent motion dynamics.}
	\vspace{0.1in}
	\label{fig:supp_ablation2}
\end{figure*}

\vspace{-0.1in}
\paragraph{Ablation on training strategies:} We further conduct an ablation study on the training strategies of \name in Fig.~\ref{fig:supp_ablation2}.
Without velocity regularization, the model predicts false-positive velocities in static regions, leading to artifacts in the rendered frames.
Without velocity re-weighting, the rendering quality degrades, resulting in blurry and inconsistent renderings.

\subsection{Qualitative Comparison to DrivingRecon~\cite{lu2024drivingrecon}}

\begin{figure*}[htbp!]
	\centering
	\vspace{0.1in}
	\includegraphics[width=\textwidth]{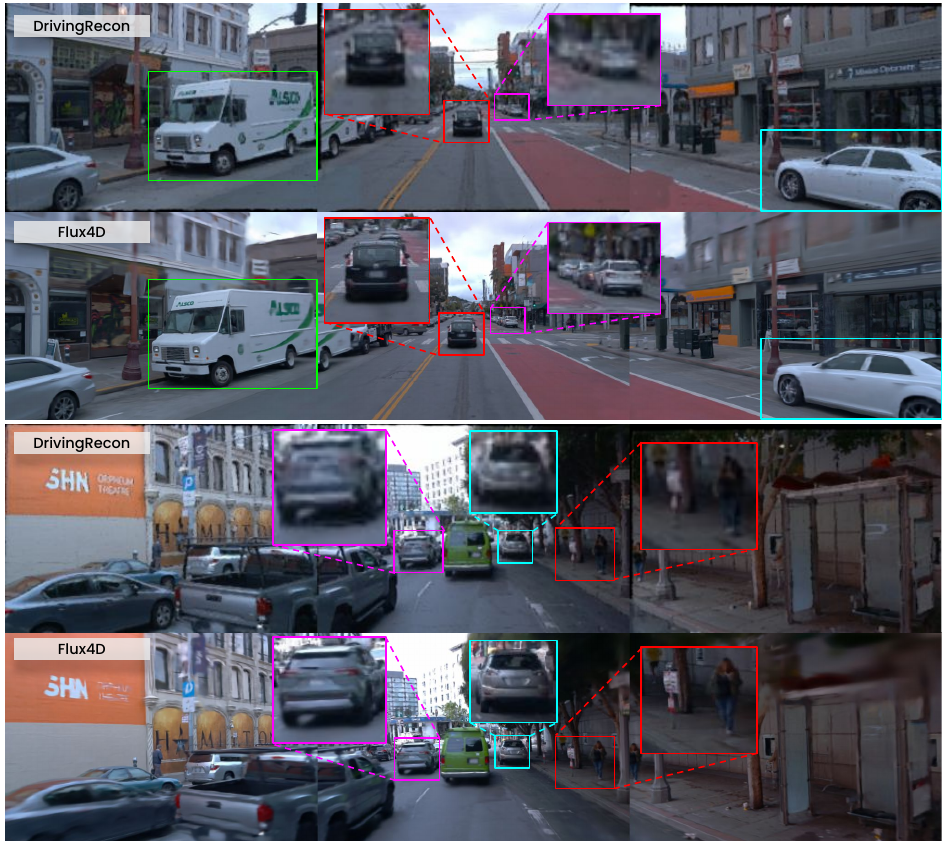}
	\vspace{-0.1in}
	\caption{\textbf{Qualitative comparison with DrivingRecon~\cite{lu2024drivingrecon} on WOD.}
	}
	\label{fig:waymo_drivingrecon}
\end{figure*}

We confirm that our evaluation setup aligns with the NVS setting in~\cite{lu2024drivingrecon} and sincerely thank the authors for in-depth discussions and for providing visual examples for comparison. We provide qualitative comparisons with DrivingRecon~\cite{lu2024drivingrecon} in Fig.~\ref{fig:waymo_drivingrecon}.
Overall, our method achieves superior reconstruction quality, operating at a higher resolution while avoiding artifacts present in DrivingRecon. However, we acknowledge that DrivingRecon produces better background reconstruction (\textit{e.g}., sky, buildings), potentially due to its pixel-aligned Gaussian prediction. For simplicity, we currently sample points in a spherical image and use the same network for both foreground and background, which may lead to suboptimal results. We leave more advanced background modeling for future work (Sec.~\ref{sec:supp_limitations}).

\subsection{Future Prediction with \textbf{\name{}}}

\begin{figure*}[htbp!]
	\centering
	\includegraphics[width=\textwidth]{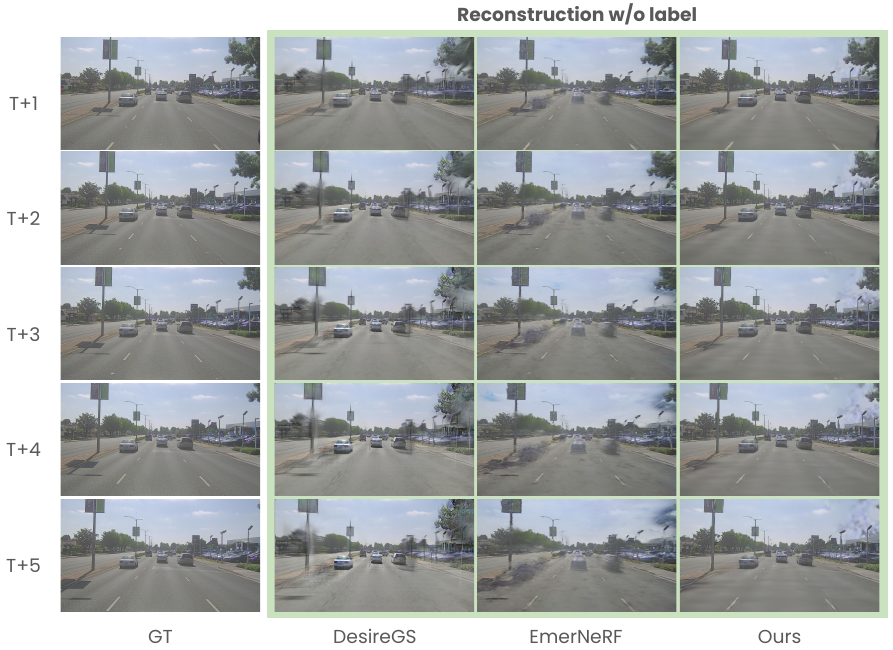}
	\vspace{-0.1in}
	\caption{\textbf{Comparison against unsupervised reconstruction methods on future prediction.}}
	\label{fig:supp_future_pred}
\end{figure*}

\begin{figure*}[t]
	\centering
	\includegraphics[width=1.0\textwidth]{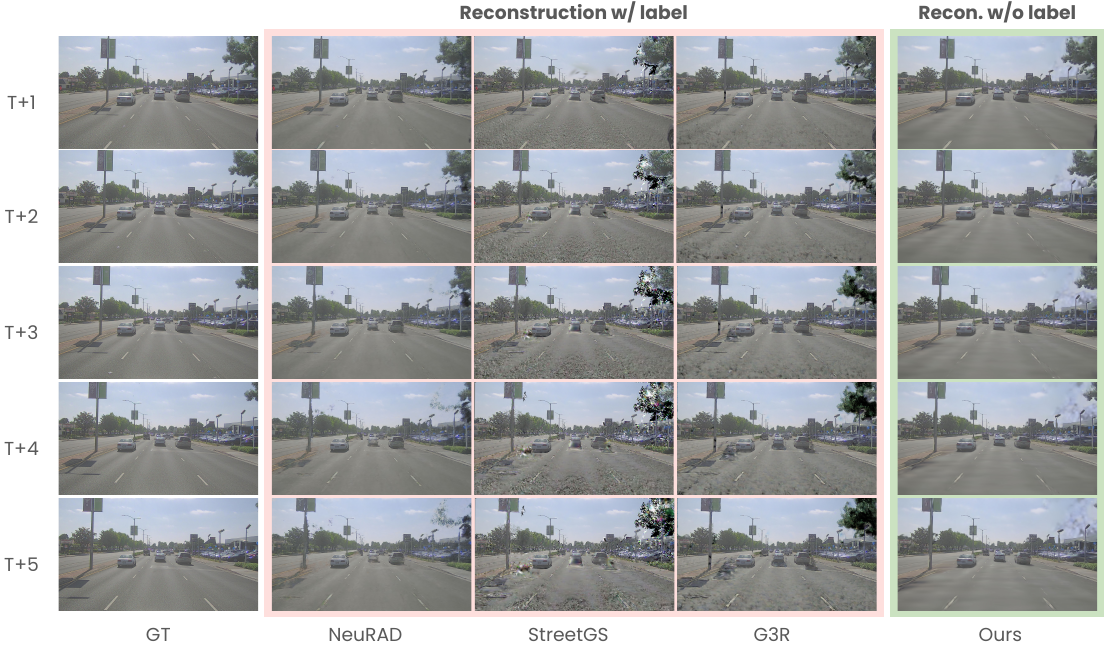}
	\caption{\textbf{\name achieves competitive performance on future frames, even compared to methods that reconstruct with labels.} Unlike these methods, which rely on labels to render extrapolation views, \name learns to predict future dynamics in an unsupervised manner, highlighting its potential for scene understanding and forecasting.}
	\label{fig:supp_future_pred2}
	\vspace{-0.1in}
\end{figure*}

\name achieves state-of-the-art performance on future prediction compared to other unsupervised reconstruction methods. We share 
additional qualitative results in Fig. \ref{fig:supp_future_pred}. Additionally, Fig. \ref{fig:supp_future_pred2} showcases \name's competitive 
qualitative results even when compared with methods that rely on labels for rendering extrapolation views.

\begin{figure*}[t]
	\centering
	\includegraphics[width=1.0\textwidth]{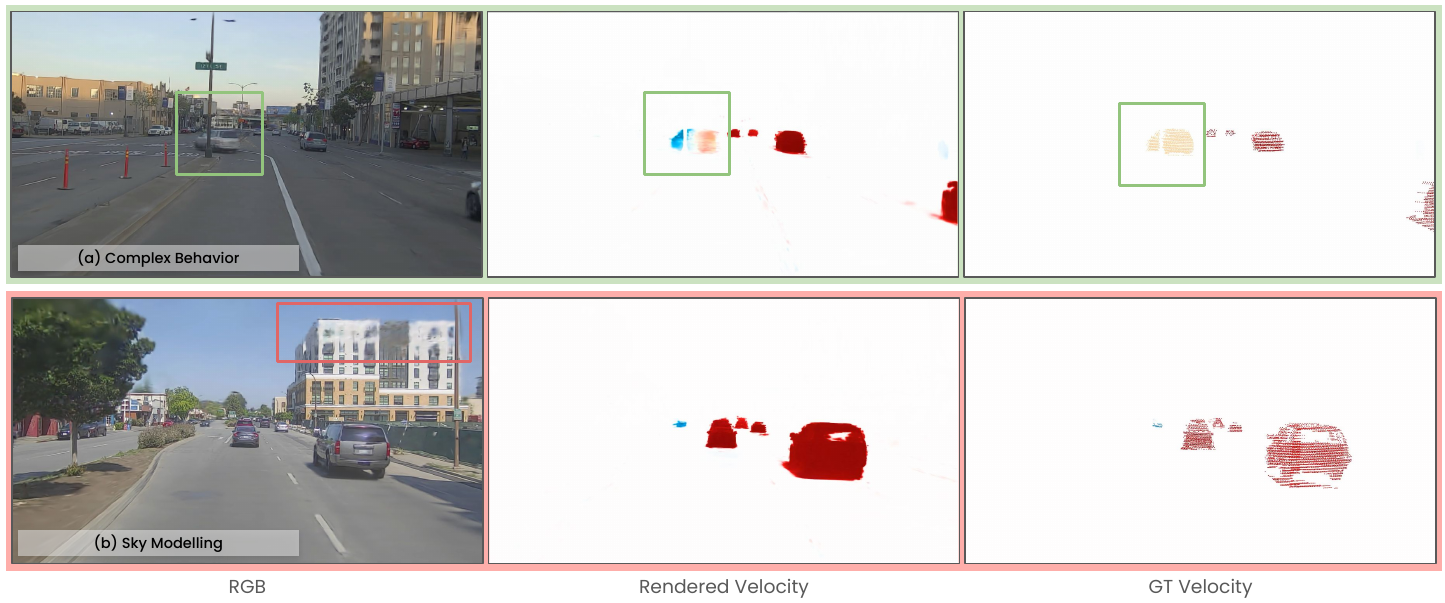}
	\vspace{-0.2in}
	\caption{\textbf{Artifacts and limitations of \name.}  
		(a) \name may struggle to accurately decompose the motion of actors exhibiting complex behaviors, such as abrupt left turns, particularly in occluded scenarios. Scaling to more diverse and larger-scale scenarios could help mitigate this issue.  
		(b) Distant regions may appear blurry due to missing LiDAR points and our simplified sky modeling approach, which represents the sky using randomly placed 3D Gaussians on a single spherical plane. Incorporating more advanced sky modeling techniques~\cite{unisim} and denser point initialization (\textit{e.g}., structure from motion) could further improve performance.}
	\label{fig:limitation}
	\vspace{0.1in}
\end{figure*}

\subsection{Additional Qualitative Examples}

\begin{figure*}[t]
	\centering
	\includegraphics[width=1.0\textwidth]{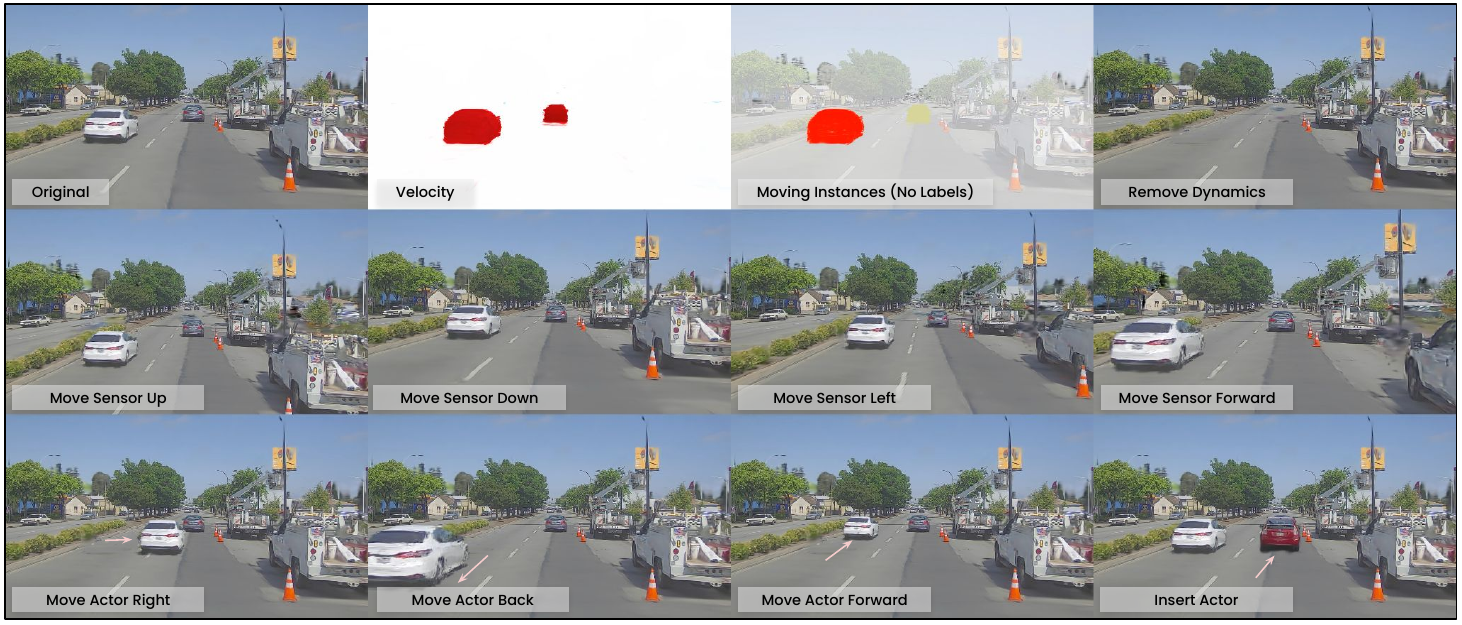}
	\caption{\textbf{Realistic and controllable simulation with \name (Part 1/2).} 
Our approach decomposes 4D dynamic scenes with accurate motion flows, enabling precise control over both sensor viewpoints and dynamic actor placements. We demonstrate various scene modifications, including novel sensor placements, actor removal, insertion and manipulation. Importantly, we achieve this without reliance on any labels or annotations.
		}
	\label{fig:supp_sim}
\end{figure*}

\begin{figure*}[t]
	\centering
	\includegraphics[width=1.0\textwidth]{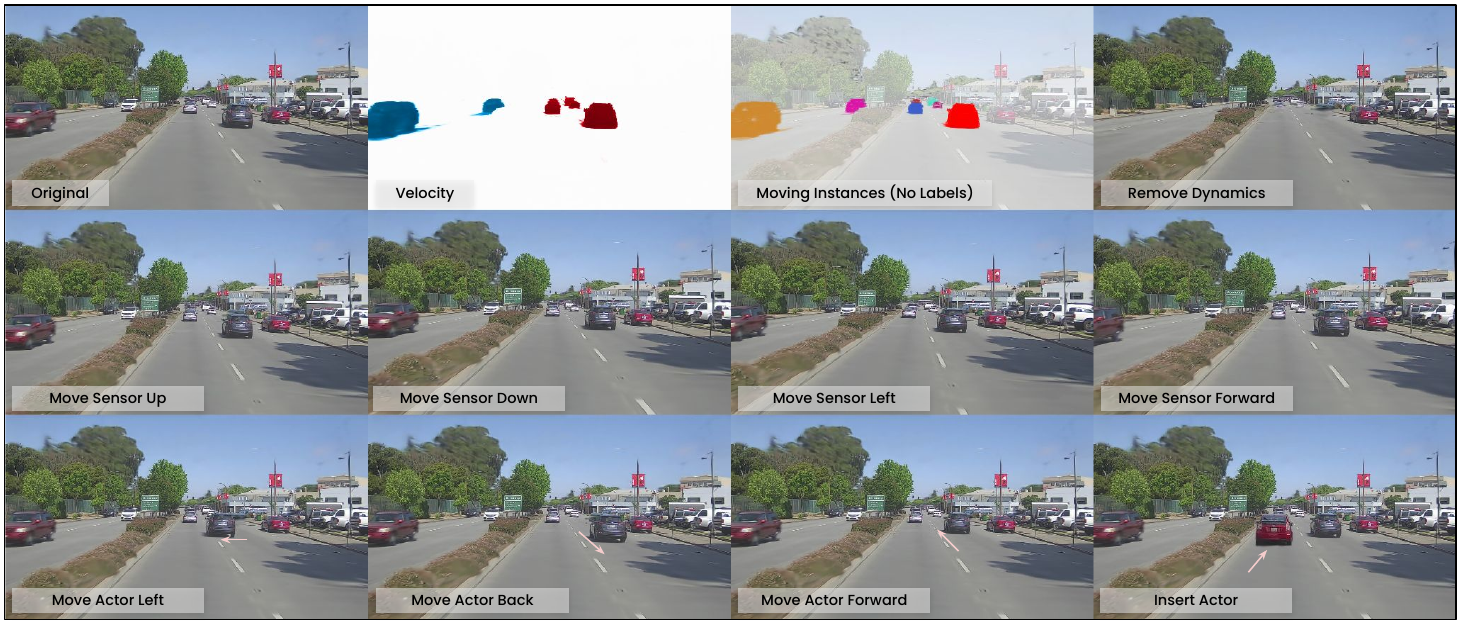}
	\includegraphics[width=1.0\textwidth]{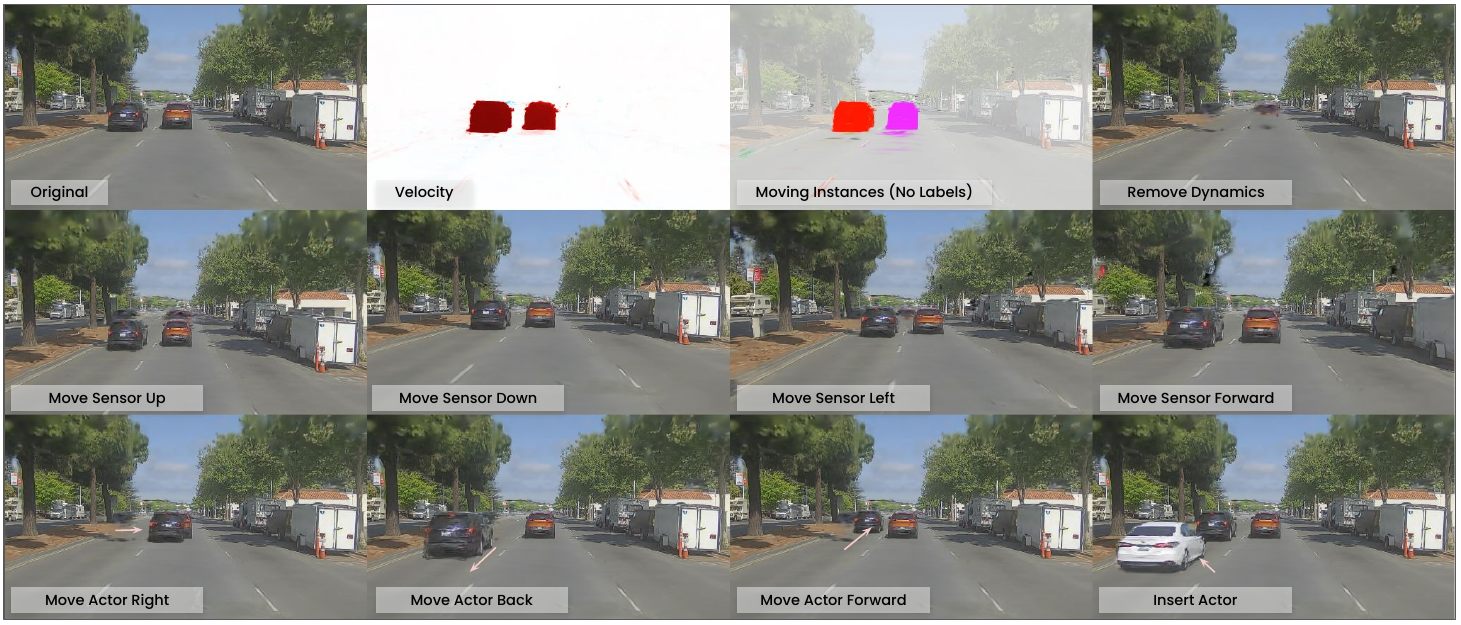}
	\includegraphics[width=1.0\textwidth]{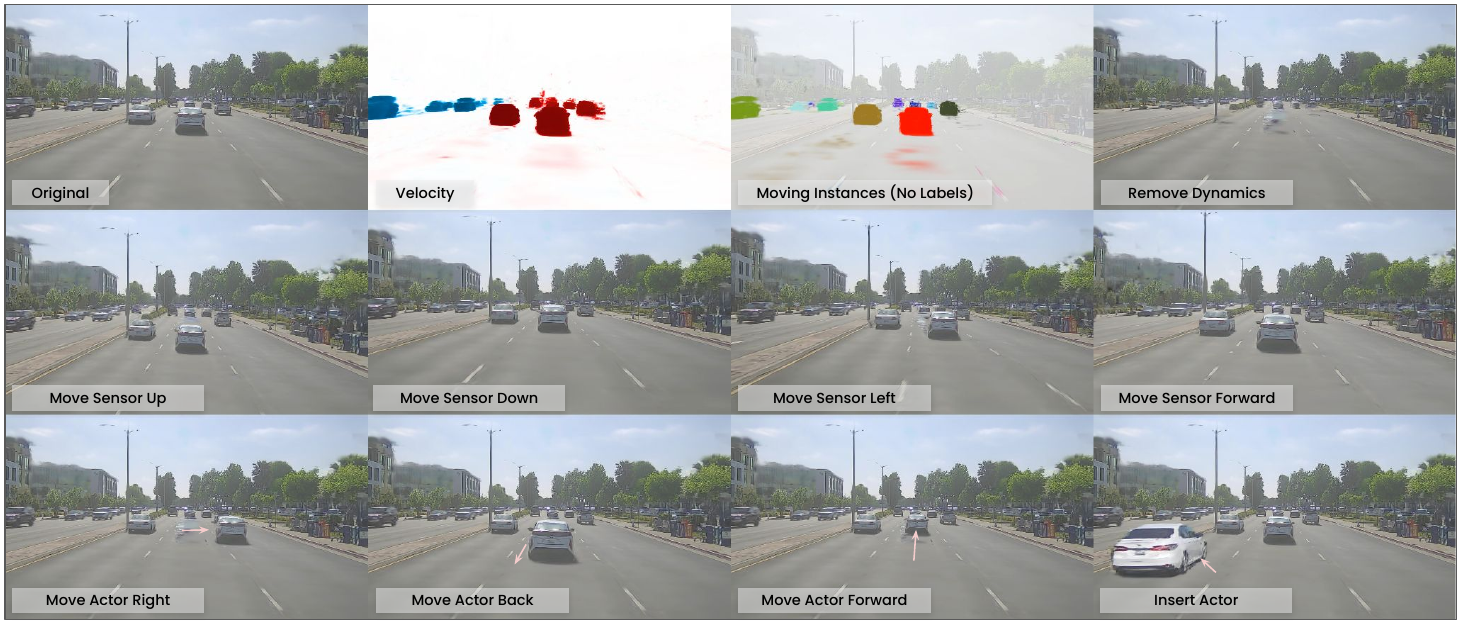}
	\vspace{-0.25in}
	\caption{\textbf{Realistic and controllable simulation with \name (Part 2/2).}
	}
	\label{fig:supp_sim2}
\end{figure*}

\paragraph{Controllable simulation:} We provide additional examples of realistic and controllable simulation with \name in Fig.~\ref{fig:supp_sim} and Fig.~\ref{fig:supp_sim2}.
Moving objects are clustered into instances using DBSCAN~\cite{ester1996density}, which can be manipulated to create diverse and counterfactual scenarios. We additionally show \name's ability to handle various 
camera movement and actor manipulation. 

\begin{figure*}[t]
	\centering
	\includegraphics[width=1.0\textwidth]{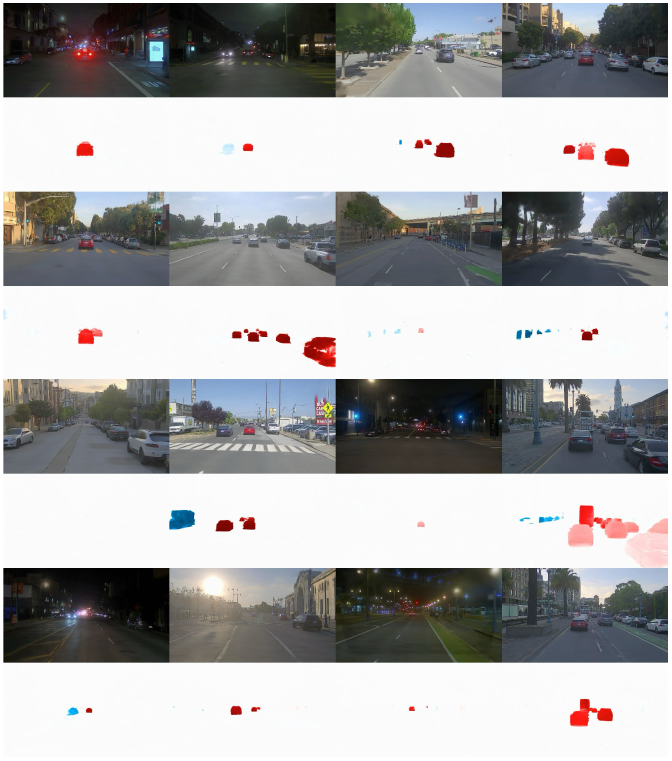}
	\vspace{-0.15in}
	\caption{\textbf{Reconstructing diverse 4D urban driving scenes at scale with \name (Part 1/2).}}
	\label{fig:supp_diversity}
\end{figure*}

\begin{figure*}[t]
	\centering
	\includegraphics[width=1.0\textwidth]{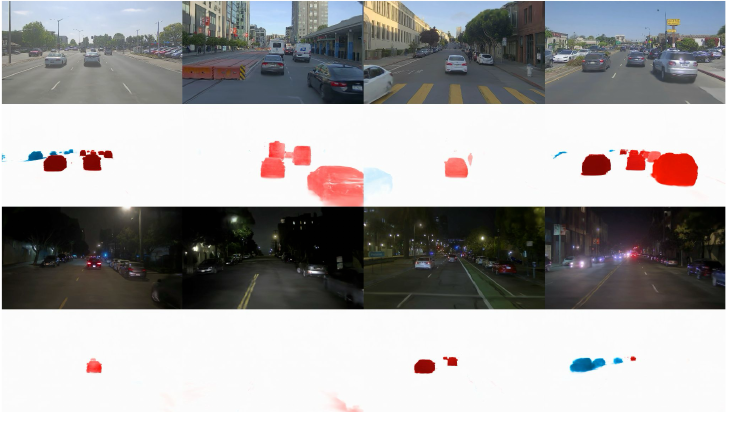}
	\vspace{-0.15in}
	\caption{\textbf{Reconstructing diverse 4D urban driving scenes at scale with \name (Part 2/2).}}
	\label{fig:supp_diversity2}
\end{figure*}

\vspace{-0.15in}
\paragraph{Generalization to diverse scenes:} As an unsupervised reconstruction method at heart, 
\name excels at reconstructing diverse 4D urban driving scenes at scale without relying on any annotations or pretrained vision models. We showcase additional examples in Fig.~\ref{fig:supp_diversity} and Fig.~\ref{fig:supp_diversity2}.

\section{Discussions}
\label{sec:discussion}

We would like to highlight the main differences between our method and several related works. We hope this disccusion can help clarify the novelty and contirbutions of our \name.

\subsection{Differences to G3R} 
We bring the iterative refinement idea in G3R~\cite{chen2025g3r} to further improve the reconstruction realism of \textit{Flux4D-base}. Our approach differs in several key aspects. We omit the neural Gaussian representation and the flatness regularization term. Thanks to our scalable unsupervised training and high-quality motion estimation, we achieve superior performance as indicated in Table~\ref{tab:comparison} (interpolation setting), \ref{tab:comparison_full_snippet} (full sequence setting), and \ref{tab:future_prediction} (extrapolation setting) of the main paper. Moreover, our method only require $N=3$ iteration compared to G3R's $N=24$ iterations to achieve competitive rendering quality, resulting in significantly faster reconstruction than G3R (3.9s v.s. 17s, Table~\ref{tab:comparison}). 

\subsection{Comparison to DrivingRecon and STORM}	
Most recently, DrivingRecon~\cite{lu2024drivingrecon} and STORM~\cite{yang2025storm} explore similar problem of unsupervised generalizable 4D reconstruction for driving scenes, using feed-forward networks to predict the velocities of 3D Gaussians. We highlight three important distinctions as follows:

\vspace{-0.1in}
\paragraph{Model dependency:} DrivingRecon and STORM rely on pre-trained vision models (\textit{i.e.}, DeepLabv3+, SAM, ViT-Adapter). In contrast, Flux4D is fully unsupervised and does not depend on any pretrained vision models or foundational priors. Despite its simpler design, \textit{Flux4D} achieves superior performance against DrivingRecon (Table~\ref{tab:drivingrecon} in the main paper).
	
\vspace{-0.1in}
\paragraph{Resolution and number of views:} While DrivingRecon and STORM operate on low-resolution inputs ($256 \times 512$ and $160 \times 240$ respectively) with limited views (3-4 frames), \textit{Flux4D} supports high-resolution (1080x1920) inputs with 20+ frames, enabling significantly more detailed reconstructions.

\vspace{-0.1in}
\paragraph{Efficiency:} \textit{Flux4D} is more computationally efficient and scalable thanks to our minimalist design. In contrast to DrivingRecon and STORM, which require approximately 480 and 768 A100 GPU hours (confirmed by authors) respectively, \textit{Flux4D} trains with only 140 L40S GPU hours.

We provide a detailed summary of the differences between \name and these two methods in Table~\ref{tab:concurrent}.

\begin{table}[htbp!]
	\centering
	\caption{\textbf{High-level comparison of \name with DrivingRecon~\cite{lu2024drivingrecon} and STORM~\cite{yang2025storm}.}}
	\label{tab:concurrent}
	\resizebox{\textwidth}{!}{
	\begin{tabular}{lccc}
		\toprule
		\textbf{Method} & \textbf{DrivingRecon} & \textbf{STORM} & \textbf{\name} \\ \midrule
		Image resolution & $256\times512$ & $160\times240$ & $\geq 1080\times1920$ \\
		Training GPU hours & $\sim$480 (A100 80GB) & $\sim$768 (A100 80GB) & $\sim$140 (L40S 48GB) \\ 
		Support LiDAR input & No & No & Yes \\
		Usage of Pre-trained vision model & Yes & Yes & No \\
		Number of frames (3 cameras) & $3$ & $4$ & $20+$ \\
		\bottomrule
	\end{tabular}
}
	\vspace{-0.2in}
\end{table}

\section{Limitations}
\label{sec:supp_limitations}
\subsection{Artifacts and Potential Enhancements}

\paragraph{Complex behaviors:}
In Fig.~\ref{fig:limitation} (a), \name{} fails to accurately predict the flow of the left-turning vehicle due to occlusions from the 
pole. \name{} arrives at the nonrigid solution for the vehicle by predicting 2 clusters of opposite flows. 
Potential enhancements include enforcing rigidity for moving instances or augmenting \name with memory. 

\vspace{-0.1in}
\paragraph{Faraway regions:} \name initializes from LiDAR point cloud. However, due to the sparsity of LiDAR data, 
some faraway regions such as trees and sky are often not covered as shown in Fig.~\ref{fig:limitation} (b).
Potential enhancements include learning multi-layer high-resolution skyboxes or panorama images or adding density control strategy.  

\subsection{Limitations and Future Work}
\paragraph{Complex motion pattern:} Despite \name{}'s significant performance improvements over prior work, accurately reconstructing flow for highly dynamic actors or those with complex motion patterns remains challenging. Scaling to larger, more diverse datasets and adopting more advanced velocity models or implicit flow representations could help mitigate this limitation.

\vspace{-0.1in}
\paragraph{Requirement of poses:} Besides, \name requires posed cameras and LiDAR initialization, limiting its applicability to scenarios with unknown camera parameters. Integrating pose-free methods like DUSt3R~\cite{wang2024dust3r} would eliminate the dependency on camera poses, enabling reconstruction from casual video captures and increasing Flux4D's versatility to uncalibrated or partially calibrated sensor setups.

Overall, we hope that our simple and scalable design serves as a foundation for the community to build upon, enabling further advancements in 4D reconstruction.

\section{Broader Impact}
\label{sec:broader_impact}
\name is an unsupervised 4D reconstruction method that can be applied to various applications, including autonomous driving, robotics, and augmented reality. Especially for autonomous driving, its ability to accurately model the highly dynamic nature of the urban environemnt, coupling with its unsupervised nature, allow it to be a powerful pretraining method for various downstream tasks such as motion forecasting, object labelings, and simulator training. Additionally, with more expressive motion modeling, \name could be useful to other domain such as augmented reality or robotics, where accurate 4D reconstruction is crucial for scene understanding and interaction.
Creating digital twins of real-world locations may raise privacy concerns. Additionally, our system may exhibit unstable performance or unintended behavior on different datasets, particularly when sensory data are sparse or noisy.

\begin{table}[htbp!]
	\centering
	\resizebox{\textwidth}{!}{
		\begin{tabular}{l c l}
			\toprule
			\textbf{Experiment} & \textbf{L40S Hours} & \textbf{Comments} \\ \midrule
			Table~\ref{tab:comparison} \& \ref{tab:comparison_full_snippet} (Interpolation \& Full Seq) & $\sim$ 400 & We use the same \name model for both \\
			Table~\ref{tab:drivingrecon} (DrivingRecon) & $\sim$ 150 & No baseline training needed \\
			Table~\ref{tab:future_prediction} (Future Prediction) & $\sim$ 300 & No training for Depthsplat, L4GM \\
			Table~\ref{tab:ablation} \& \ref{tab:ablation_loss} (Ablation) & $\sim$ 900 & This requires 5 more \name's training jobs \\
			Fig.~\ref{fig:argoverse_waymo} (Argoverse2 \& WOD) & $\sim$ 300 & WOD: 150hr, Argoverse2: 150hr \\
			Fig.~\ref{fig:scaling_laws} (Scaling Analysis) & $\sim$ 1200 & Pandaset: 750hr, WOD: 450hr \\
			Others (data generation \& demos) & $\sim$ 10 & No data generation needed, estimated 10hr for all demos \\ 
			\bottomrule
		\end{tabular}
	}
	\vspace{0.05in}
	\caption{Summary of GPU hours used for the final experiments.}
	\label{tab:gpu_hours}
	\vspace{-0.2in}
\end{table}

\section{Computation Resources}
\label{sec:computation}

We provide the estimated GPU usage for the final experiments in Table~\ref{tab:gpu_hours}, where all hours are converted to 1xL40S hours.

\section{Licenses of Assets}
\label{sec:license}
We summarize the licenses and terms of use for all assets (datasets, software, code) in Table~\ref{tab:license}. We use the Waymo Open Dataset (WOD) only for research purposes and solely for final benchmarking, in accordance with its license requirements.

\begin{table}[htbp!]
	\centering
		\resizebox{0.8\textwidth}{!}{
		\begin{tabular}{l l l}
			\toprule
			\textbf{Assets} & \textbf{License} & \textbf{URL} \\ 
			\midrule
			\textit{Datasets} \\
			PandaSet~\cite{xiao2021pandaset} & CC BY 4.0 & \url{https://scale.com/open-av-datasets/pandaset} \\
			WOD~\cite{waymo} & Custom\tablefootnote{\url{https://waymo.com/open/terms}} & \url{https://waymo.com/open/} \\
			ArgoVerse~\cite{wilson2023argoverse} & CC BY-NC-SA 4.0 & \url{https://www.argoverse.org/av2.html}\\
			\midrule
			\textit{Codebases} \\
			PyTorch~\cite{paszke2019pytorch} & Custom\tablefootnote{\url{https://github.com/pytorch/pytorch/blob/main/LICENSE}} & \url{https://github.com/pytorch/pytorch} \\
			Torchsparse~\cite{tangandyang2023torchsparse} & MIT & \url{https://github.com/mit-han-lab/torchsparse} \\
			gsplat~\cite{gsplat} & Apache-2.0 & \url{https://github.com/nerfstudio-project/gsplat} \\
			Omnire~\cite{chen2024omnire} & MIT & \url{https://github.com/ziyc/drivestudio} \\
			\midrule 
			\multicolumn{3}{l}{\textit{Baseline Codebases (Comparisons purposes only)}} \\
			NeuRAD~\cite{tonderski2024neurad} & Apache-2.0 & \url{https://github.com/georghess/neurad-studio} \\
			Depthsplat~\cite{xu2024depthsplat} & MIT & \url{https://github.com/cvg/depthsplat} \\
			EmerNeRF~\cite{yang2023emernerf} & Custom\tablefootnote{\url{https://github.com/NVlabs/EmerNeRF/blob/main/LICENSE}} & \url{https://github.com/NVlabs/EmerNeRF} \\
			L4GM~\cite{ren2025l4gm} & Apache-2.0 & \url{https://github.com/nv-tlabs/L4GM-official} \\
			DeSiRe-GS~\cite{peng2024desire} & N/A & \url{https://github.com/chengweialan/desire-gs} \\
			\bottomrule
		\end{tabular}
	}
	\vspace{0.05in}
	\caption{Summary of the licenses of assets.}
	\label{tab:license}
	\vspace{-0.2in}
\end{table}

\clearpage  %
\FloatBarrier

\end{document}